%% file: main.tex
\pdfoutput=1
\documentclass{article}

\usepackage{iclr2027_conference,times}
\iclrfinalcopy

\input{preamble}

\input{sections/title}

\begin{document}

\maketitle
\lhead{Preprint}

\input{sections/abstract}

\section{Introduction}
\vspace{-5pt}
\label{sec:introduction}
\input{sections/intro}

\section{Benchmark Construction}

\subsection{Data}
\label{sec:behavioral-data}
\vspace{-5pt}
\input{sections/behavioral-data}

\subsection{Evaluation Tasks}
\label{sec:evaluation-tasks}
\vspace{-5pt}
\input{sections/behaviorbench/evaluation-tasks}

\subsubsection{Behavior Prediction and Simulation}
\vspace{-5pt}
\label{sec:task-predict-simulate}
\input{sections/behaviorbench/predict-simulate}

\input{sections/behaviorbench/predict-simulate-econ}

\input{sections/behaviorbench/predict-simulate-bigfive}

\subsubsection{Strategic Decision-Making}
\vspace{-5pt}
\label{sec:task-strategic}
\input{sections/behaviorbench/strategic}

\subsubsection{Subject-trait Inference}
\label{sec:task-infer}
\vspace{-5pt}
\input{sections/behaviorbench/infer-traits}

\subsubsection{Behavioral Knowledge Application}
\label{sec:task-knowledge}
\vspace{-5pt}
\input{sections/behaviorbench/apply-knowledge}

\subsection{Metrics}
\label{sec:eval-metrics}
\input{sections/benchmarking/eval-metrics}

\vspace{-5pt}
\subsection{Comparison with Existing Benchmarks}
\vspace{-5pt}
\label{sec:benchmark-comparison}
\input{sections/behaviorbench/comparison}

\vspace{-3pt}
\section{Benchmarking Foundation Models}
\vspace{-5pt}
\label{sec:models}

\input{tables/main-results}
\subsection{Models to Evaluate}
\vspace{-5pt}
\label{sec:models-baselines}
\input{sections/benchmarking/baselines}

\subsection{\bench Leaderboards}
\vspace{-5pt}
\label{sec:eval-setup}
\input{sections/behaviorbench/eval-setup}

\section{Results}
\label{sec:results}
\vspace{-5pt}

\subsection{Model Capabilities}
\vspace{-5pt}
\label{sec:results-capabilities}
\input{sections/results/capabilities}

\needspace{20\baselineskip}
\section{Co-Optimize Individual-level and Distributional Performance}
\vspace{-5pt}
\label{sec:discussion-generalization}
\input{sections/discussion/generalization-summary}

\section{Conclusion}
\vspace{-5pt}
\label{sec:conclusion}
\input{sections/conclusion}

\bibliographystyle{iclr2027_conference}
\bibliography{references}

\clearpage
\appendix

\section{Related Work}
\label{app:related}

\subsection{Behavioral Models and Foundation Models for Behavioral Science}
\input{sections/related/foundation-models}

\subsection{AI Behavioral Science}
\input{sections/related/ai-behavioral-science}

\subsection{Benchmarks for Behavior Modeling and Economic Reasoning}
\input{sections/related/benchmarks}

\input{sections/appendix/data-details}

\input{sections/appendix/metrics-details}

\section{Model Fine-tuning}
\label{app:models-befm}
\label{sec:models-befm}
\input{sections/befm-model}

\input{sections/appendix/direct-elicitation}
\input{sections/appendix/per-task-results}
\input{sections/appendix/prompt-templates}

\input{sections/appendix/asset-licenses}

\end{document}

%% file: preamble.tex
\usepackage[utf8]{inputenc}
\usepackage[T1]{fontenc}
\usepackage{hyperref}
\usepackage{url}
\usepackage{booktabs}
\usepackage{amssymb}
\usepackage{microtype}
\usepackage{xcolor}
\usepackage{multirow}
\usepackage{makecell}
\usepackage{amsmath}
\usepackage{subcaption}
\usepackage{graphicx}
\usepackage{wrapfig}
\usepackage{xspace}
\usepackage[inline]{enumitem}
\usepackage{arydshln}
\usepackage{mdframed}

\newcommand{\modelfamily}{\texttt{Be.FM}\xspace}
\newcommand{\bench}{\texttt{BehaviorBench}\xspace}

%% file: sections/title.tex
\title{\bench: Benchmarking Foundation Models for Behavioral Science Tasks}

\author{%
Jin Huang$^{1}$\thanks{Equal contribution.} \quad
Yutong Xie$^{1}$\footnotemark[\value{footnote}] \quad
Wanli Song$^1$ \quad
Xingjian Zhang$^1$ \quad
Owen Yang$^2$ \\[2pt]
\bfseries Walter Yuan$^3$ \quad Matthew O. Jackson$^{4,5}$ \quad Qiaozhu Mei$^1$ \\[6pt]
\normalfont $^1$University of Michigan \quad $^2$Henry M. Gunn High School \quad $^3$MobLab \\
\normalfont $^4$Stanford University \quad $^5$Santa Fe Institute
}

%% file: sections/abstract.tex
\begin{abstract}
Foundation models have been increasingly applied to behavioral science domains such as psychology, sociology, and economics. While these models show promise in tasks such as survey response prediction and human-subject experiment simulation, there remains no systematic understanding of how well they perform across diverse behavioral science tasks. We introduce \bench, a comprehensive benchmark that evaluates foundation models along four core capabilities: (1) behavior prediction and simulation, (2) strategic decision-making, (3) subject-trait inference, and (4) behavioral knowledge application. Crucially, \bench evaluates model outputs at both the individual and distributional levels, capturing not only per-subject accuracy but also population-level alignment, an essential requirement for behavioral validity.
Our evaluation shows that \bench remains challenging for leading general-purpose LLMs and behavior foundation models that are specifically trained with behavioral data.
We find that individual-level and distributional performance do not always align. General-purpose LLMs tend to underestimate the diversity of human responses, whereas behavior foundation models often lag behind at individual-level prediction. Our investigation further demonstrates how fine-tuning on diverse behavioral data can improve both individual-level prediction and distributional alignment, balancing these two objectives.
Our results highlight the importance of evaluation at both individual and distributional levels, establishing \bench as a foundation for developing and assessing behaviorally aligned AI systems.
Our \bench and models can be accessed via \url{https://umich-foreseer.github.io/behaviorbench/}.
\end{abstract}

%% file: sections/intro.tex
\input{sections/figures/ind-vs-dist}

Foundation models have transformed many fields. Trained on vast and diverse data, they can adapt across tasks and are increasingly used not only for prediction, but also for reasoning, simulation, and scientific discovery~\citep{bommasani2021foundation,park2023generative,wang2023scientific}.
One field with particularly transformative potential is \emph{behavioral science}: the science of understanding, predicting, and shaping human behaviors.
Drawing on multiple disciplines,
including psychology, sociology, anthropology, and economics, behavioral science provides foundations for understanding decision-making dynamics at individual, group, and societal levels~\citep{gintis2014bounds}.

Recent work has started to apply general-purpose foundation models, including commercial and open-source large language models, to behavioral science \citep{demszky2023using,meng2024ai,bail2024can}. These models show promise in improving survey research \citep{argyle2023out,grossmann2023ai}, supporting agent-based modeling \citep{park2023generative,park2024generative}, and predicting human-subject experiment results \citep{hewitt2024predicting,lippert2024can}.
In parallel, researchers are also developing behavioral-science-specific foundation models by fine-tuning LLMs on behavior-related data. These models have demonstrated improved performance in simulating human behavior,
predicting survey responses, and supporting behavioral research workflows~\citep{BeFM-1,centaur,socsci210}.

Despite this growing body of work, we still lack a systematic understanding of how well foundation models perform across diverse behavioral science tasks, contexts, and populations.
Existing behavior benchmarks are often limited by a narrow task family, such as survey response prediction or behavior simulation~\citep{OdysSim,simbench,subpop,sociobench}.
As a result, models' capabilities in important behavior science tasks, such as inferring latent traits of human subjects or applying behavioral knowledge to scientific research, remain unmeasured.

To address this limitation, we introduce \bench, a comprehensive benchmark for systematically evaluating foundation models on behavioral science tasks.
We organize our evaluation around the observation that human behavior is shaped by multiple factors like the behavioral context, the subject's traits, and underlying motivations~\citep{BeFM-1}. Formally, human behavior can be represented as a conditional probability
\vspace{-5pt}
\begin{equation}
p(y\mid x, c; \mathbb{K}),
\label{eqn:framework}
\end{equation}
where $y$ denotes a behavior, $x$ the characteristics of the human subject, $c$ the contextual variables defining the environment or intervention, and $\mathbb{K}$ represents relevant behavioral knowledge.
With the majority of existing benchmarks focusing only on the prediction of $y$, however, to best support behavioral science, a behavior foundation model should support multiple types of inference within this framework. As a consequence, \bench evaluates models along four core capabilities: (1) behavior prediction and simulation (forward inference over $y$ to mimic a human subject), (2) strategic decision-making (the generation of $y$ to achieve a specified goal), (3) subject-trait inference (inverse inference over $x$), and (4) the application of behavioral science knowledge ($\mathbb{K}$) to research problems.
Built on a diverse collection of datasets,
\bench comprises 12 distinct tasks that span these four capabilities.

In addition, for central behavioral tasks such as behavior simulation, researchers care about the accuracy at two complementary levels: \emph{individual level}, where the objective is to predict individual outcomes; and \emph{distributional level}, where the objective is to predict the outcome distribution of the population~\citep{xie2025distributional}. However, most existing benchmarks emphasize only one of these levels~\citep{OdysSim, simbench, sociobench} (Table~\ref{tab:benchmark-comparison}).
In contrast, our \bench provides dedicated evaluations at both the \emph{individual} and the \emph{distributional} levels.

Our main contributions are:
\begin{itemize}[leftmargin=*,itemsep=0pt,topsep=2pt]
    \item We introduce \bench, a comprehensive benchmark covering 12 behavioral science tasks across four capabilities, with evaluation at both the individual and distributional levels.

    \item We systematically evaluate 20 LLMs, including general-purpose frontier LLMs, and behavior foundation models that are specifically trained on behavioral data. Our results show that \bench is challenging for all models evaluated, with Gemini 3.1 Pro and Claude Opus 5 standing out on the leaderboards.

    \item We find that individual-level and distributional performance do not always align.
    In particular, general-purpose LLMs tend to underestimate the diversity of human responses, despite more accurately predicting the population mean, whereas behavior foundation models better preserve response diversity.

    \item Our further investigation into fine-tuning LLMs with diverse behavioral data shows the possibility of simultaneously improving the individual-level and distributional performance, demonstrating the pathway to behaviorally aligned AI systems.

\end{itemize}

%% file: sections/figures/ind-vs-dist.tex
\begin{figure}[!b]
    \centering
    \begin{minipage}[c]{0.44\textwidth}
        \centering
        \includegraphics[width=\linewidth]{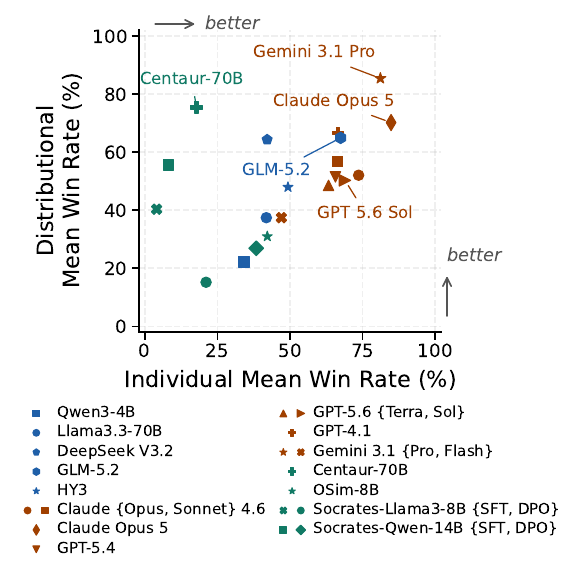}
    \end{minipage}\hfill
    \begin{minipage}[c]{0.54\textwidth}
        \centering
        \includegraphics[width=\linewidth]{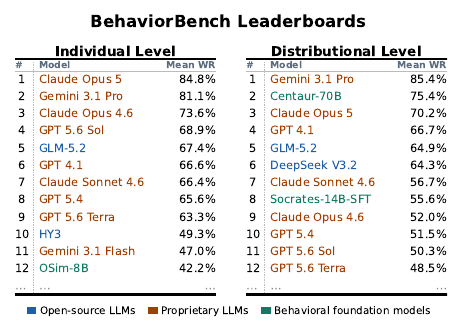}
    \end{minipage}
    \vspace{-5pt}
    \caption{Aggregated evaluation results of foundation models on \bench.
    \emph{Left:} individual- vs.\ distributional-level performance in mean win rate (\%) across 20 models.
    \emph{Right:} Snapshots of the top 12 models on the BehaviorBench leaderboards, with models ranked at the individual and distributional levels by mean win rate (Mean WR, \%).
    }
    \label{fig:ind-vs-dist}
\end{figure}

%% file: sections/behavioral-data.tex
Capturing the breadth of human behavior demands data of comparable breadth.
Following~\citet{BeFM-1}, \bench draws on three sources of behavior data to construct the evaluation. Details are in Appendix \ref{app:data-details}.

\begin{enumerate}[leftmargin=15pt]
    \item \textbf{Experimental data. }
Experimental records capture behavioral patterns of diverse subjects in controlled environments, offering key insights into individual and group decision-making. We consider a collection of experiments from MobLab\footnote{MobLab: \url{https://www.moblab.com/}.}, which covers a diverse population of human subjects and their play records in multiple economics games~\citep{mei2024turing}.
We choose this dataset because it is one of the largest collections of controlled economic-game records, covering tens of thousands of subjects from more than 50 countries.
\bench uses an evaluation subsample of 12,685 behavioral observations from 6,279 unique subjects.
\item \textbf{Survey data. }
Survey data provide self-reported insights into personality traits, attitudes, and preferences. We use the publicly available Big Five Personality Test dataset\footnote{\raggedright\url{https://www.kaggle.com/datasets/lucasgreenwell/ocean-five-factor-personality-test-responses}.} as a data source. This dataset consists of demographic information of subjects (e.g., race, age, gender), along with their answers to 50 survey questions.
The answers can be aggregated into five personality scores.
We choose this dataset because it is one of the largest publicly available personality-survey datasets, with around one million responses and respondents from 158 countries.
\bench includes an evaluation subsample of 6,000 responses from 1,886 unique subjects.
\item \textbf{Literature data. }
The behavioral science literature documents existing theories and scientific findings about human behavior. We collect titles and abstracts of articles from \emph{American Economic Review} and \emph{Nature Human Behaviour} that were published in 2025. We then follow \citet{massw} to decompose each title--abstract pair into a structured research workflow that reflects how the studies are done.
Such data are included in \bench to evaluate foundation models' ability in scientific reasoning and assisting behavioral science research.
\end{enumerate}

%% file: sections/behaviorbench/evaluation-tasks.tex
\input{tables/benchmark-task-summary}

By formulating the generation of human behavior as $p(y\mid x,c,\mathbb{K})$, \bench evaluates foundation models with four categories of behavioral tasks (Table~\ref{tab:benchmark-task-summary}): (1) behavior prediction and simulation (forward inference over $y$), (2) strategic decision-making (the generation of $y$ to achieve a goal), (3) subject-trait inference (inverse inference over $x$), and (4) the application of behavioral science knowledge ($\mathbb{K}$) for scientific reasoning.
Formally, let $\mathcal{Y}$ denote the space of human behaviors (e.g., actions in experiments and survey responses), $\mathcal{X}$ the space of human subject traits (e.g., demographics and personalities), and $\mathcal{C}$ the space of contextual variables (e.g., experiment setup and intervention).

%% file: tables/benchmark-task-summary.tex
\begin{table*}[t]
\centering
\small
\setlength{\tabcolsep}{4pt}
\def\arraystretch{1.2}
\resizebox{\textwidth}{!}{%
\begin{tabular}{p{2.7cm}p{7.5cm}ccc}
\toprule
 &  &  & \multicolumn{2}{c}{\textbf{Eval. Metric}} \\
\cmidrule(lr){4-5}
\textbf{Capability} & \textbf{Task} & \textbf{\# Eval. Data} & \textbf{Ind.} & \textbf{Dist.} \\
\midrule

\multirow{6}{2.7cm}{\textbf{Behavior prediction and simulation}}
& Single-round game behavior simulation (Game Behav.\ Sim.) & 200/game & -- & W \\
& Multi-round game behavior prediction (Multi-Round Pred.) & 500/game & MAE & W \\
& Single-round game behavior prediction given observations from other games (Across-Ctx Pred.) & 750/game & MAE & W \\
& Survey response prediction given demographics (Demo.\ To Resp.) & 1,000 & Acc & W \\
& Sequential survey response prediction (Seq.\ Resp.\ Pred.) & 1,000 & Acc & W \\
& Masked survey response prediction (Masked Resp.\ Pred.) & 1,000 & Acc & W \\
\midrule

\textbf{Strategic\quad\quad decision-making}
& Strategic game play & 1,000 & Win (\%) & -- \\
\midrule

\multirow{3}{2.7cm}{\textbf{Subject-trait inference}}
& Personality score prediction given demographics (Demo.\ To Pers.) & 1,000 & MAE & W \\
& Personality score prediction given scores from other dimensions (Across-Dim Pers.\ Pred.) & 1,000 & MAE & W \\
& Age prediction given personality scores (Pers.\ To Demo.) & 1,000 & MAE & W \\
\midrule

\multirow{2}{2.7cm}{\textbf{Knowledge application}}
& Scientific workflow prediction & 1,084 & BLEURT & -- \\
& Economics contest problem solving & 124 & Acc & -- \\
\bottomrule
\end{tabular}}
\vspace{5pt}
\caption{Evaluation tasks in \bench, by core capabilities.
\textbf{\#Eval. Data}: evaluation set sizes. Metrics: individual level (\textbf{Ind.}) and distributional level (\textbf{Dist.}, ``W'' stands for Wasserstein distance). Per-task data sources and processing procedures are detailed in Appendix~\ref{app:data-details}.
}
\label{tab:benchmark-task-summary}
\end{table*}

%% file: sections/behaviorbench/predict-simulate.tex
Predicting and simulating human behavior is at the core of behavior modeling~\citep{sociobench, centaur, socsci210}. Behavior prediction arises naturally at two levels: the individual level
and the distributional level.

At the individual level, the task is to predict the behavior of a particular subject in a given context. Formally, this can be represented as a projection
\begin{equation}
    \mathcal{X} \times \mathcal{Y} \times \mathcal{C} \to \mathcal{Y}.
    \label{eqn:predict-individual}
\end{equation}

Concretely, a model predicts a subject's behavior $y$ (e.g., their choice in the next round) from the subject's characteristics $x$, the context $c$ (e.g., experiment instructions and setup), and partial information about the subject's behavior $y'$, which can be empty.

At the distributional level, the target is to predict a behavior distribution across a population.
To obtain a behavior distribution $Y$, we run individual-level predictions
and then evaluate whether $Y$ aligns with the empirical human behavior distribution.
\bench constructs the evaluation of this capability from two data sources below.

%% file: sections/behaviorbench/predict-simulate-econ.tex
\textbf{Game play simulation and prediction. }
We define three simulation/prediction tasks. (1) \emph{Single-round game behavior simulation.} Given the game instructions, a model predicts the distribution of the population's actions in a single (first) round. The purpose is to measure whether a model distributionally aligns with the human population when playing a game. (2) \emph{Multi-round game behavior prediction.} Given the game instructions and a subject's behavior in prior rounds, a model predicts their action in the next round. Different from (1), here we condition on a subject's decision trajectory. (3) \emph{Across-context prediction.} Given a subject's behavior in other games, a model predicts the first-round behavior in a target game.

%% file: sections/behaviorbench/predict-simulate-bigfive.tex
\textbf{Survey response prediction. }
Using the Big Five responses described in Section~\ref{sec:behavioral-data}, we define three prediction tasks. (1) \emph{Survey response prediction given demographics.} Given a subject's demographics and a single survey item, a model predicts the subject's response. This task is consistent with the setting in prior works~\citep{opinionqa, socsci210}. (2) \emph{Sequential survey response prediction.} Given a subject's responses to prior items in survey order, a model sequentially predicts the response to the next item.
(3) \emph{Masked survey response prediction.} Given all other responses, a model predicts the response to one masked survey item.

%% file: sections/behaviorbench/strategic.tex
Section~\ref{sec:task-predict-simulate} concerns predicting and simulating a subject's behavior. Here, we examine a distinct task: making strategic decisions to achieve a specified goal. In this setting, actions are taken on one's own behalf, rather than mimicking a particular human subject. This capability is especially important as AI and LLM agents increasingly make autonomous decisions and interact with humans~\citep{mei2024turing,jackson2025AI}.
Formally, this task requires a model to make the projection
\begin{equation}
    \mathcal{C} \times \mathcal{Y} \to \mathcal{Y}.
    \label{eqn:strategic}
\end{equation}
Concretely, a model generates its own decision $y$ toward the specified goal from the context $c$ (e.g., experiment instructions) and other subjects' actions.
Prior work has used economic games to measure this capability~\citep{glee, econarena, GAMA, GTBENCH}. However, most prior works have LLMs play against other LLMs; few have LLMs play against human player records~\citep{glee}.

\bench uses human play records of the Beauty Contest game on MobLab to construct an evaluation. In Beauty Contest, each player picks a number in $[0, 100]$ for every round. The winner is whoever comes closest to $\frac{2}{3}$ of the group average. At the end of each round, each player gets feedback on whether they won that round.
This game involves opponent modeling and extensive reasoning~\citep{nagel1995unraveling,ho1998iterated}.

%% file: sections/behaviorbench/infer-traits.tex
In Sections~\ref{sec:task-predict-simulate} and \ref{sec:task-strategic}, we introduced tasks for predicting, simulating, and generating behavior ($y$) under various settings. Here, we introduce the inverse tasks for inferring subjects' traits ($x$).
Characterizing individual subjects and populations is a core goal of behavioral science~\citep{yang2022click, rychalska2023synerise, demszky2023using, bleidorn2019policy}.
At the individual level, the task is to infer a latent characteristic of a particular subject from their observed behavior. Formally, a foundation model is tasked with the projection
\begin{equation}
    \mathcal{Y} \times \mathcal{X} \times \mathcal{C} \to \mathcal{X}.
    \label{eqn:infer-individual}
\end{equation}

At the distributional level, we run individual-level inferences over a population
and then evaluate whether the predicted distribution aligns with the empirical human trait distribution.

We introduce three tasks. (1) \emph{Personality score prediction given demographics.} Given a subject's demographics, a model predicts their score on a Big Five personality dimension. (2) \emph{Across-dimension personality score prediction.} Given a subject's scores on four personality dimensions, a model predicts their score on the remaining dimension. (3) \emph{Age prediction given personality scores.} Given a subject's five-dimensional personality, a model predicts the subject's age.

%% file: sections/behaviorbench/apply-knowledge.tex
Beyond prediction and simulation, a behavior foundation model should be able to represent and apply behavioral science knowledge.
These tasks take the form of applying knowledge $\mathbb{K}$ to solve the input question $q$.

\textbf{Research workflow reasoning. }
Following~\citet{massw}, we decompose a paper's title and abstract into five research-workflow aspects: context, key idea, method, outcome, and projected impact. We then define five generation tasks: key idea generation, method recommendation, outcome prediction, impact prediction, and title prediction.
This group of tasks reflects models' ability to assist behavioral research workflows.

\textbf{Complex problem solving. }
To test whether the model can apply domain knowledge to unseen problems, we use multiple-choice questions from the International Economics Olympiad (IEO) contests, spanning from 2018 to 2025\footnote{International Economics Olympiad (IEO) official website: \url{https://ieo-official.org/}.}.
Model choices are compared with official solutions.

%% file: sections/benchmarking/eval-metrics.tex
For individual-level evaluation, we use:
\begin{itemize}[leftmargin=*,itemsep=0pt,topsep=2pt]
    \item \textbf{Mean absolute error (MAE)} for the prediction of numeric quantities, including game actions, trait scores, and age. We choose MAE because it preserves the ordinal structure of the response.
    \item \textbf{Accuracy} for categorical predictions, including predicting survey responses and answering IEO questions
    ~\citep{socsci210}.
    \item \textbf{Win rate} for strategic decision-making (i.e., the Beauty Contest game), defined as the fraction of sessions that a model wins over human players.
    \item \textbf{BLEURT}~\citep{bleurt} for open-ended generation in the research workflow task, following \citet{massw}.
\end{itemize}

For distribution-level evaluation, we use the \textbf{Wasserstein distance} (noted as $W$)
to compare the predicted behavior distribution against the observed human distribution. Following \citet{subpop, socsci210}, we choose $W$ because it captures both the shape and the mean of two distributions.
More details on the evaluation metrics are in Appendix~\ref{app:metrics-details}.

%% file: sections/behaviorbench/comparison.tex
\input{tables/benchmark-comparison}
Table~\ref{tab:benchmark-comparison} compares \bench with existing benchmarks for behavioral-science-related tasks. Existing benchmarks primarily target narrow task families: SOUL-Index~\citep{OdysSim}, SocioBench~\citep{sociobench}, SubPOP~\citep{subpop}, and SimBench~\citep{simbench} evaluate behavior prediction from demographics or context, while GLEE~\citep{glee}, GTBench~\citep{GTBENCH}, and GAMA-Bench~\citep{GAMA} evaluate strategic decision-making in games. These benchmarks focus on behavior $y$. \textbf{None of them evaluates the inference of subject traits $x$ or the application of behavioral knowledge $\mathbb{K}$}.
Moreover, \bench supports evaluation at both the individual and distributional levels, while previous works tend to only evaluate at one level.
We also identify several publicly available datasets that could potentially be developed into benchmarks, including OpinionQA~\citep{opinionqa}, Psych-101~\citep{centaur}, and SocSci210~\citep{socsci210}. However, these datasets likewise do not fully cover the four capability categories.

%% file: tables/benchmark-comparison.tex
\begin{table*}[t]
\centering
\small
\setlength{\tabcolsep}{3pt}
\def\arraystretch{1.15}
\resizebox{\textwidth}{!}{%
\begin{tabular}{l:cccc|cc|c}
\hline
\makecell[l]{\textbf{Capability}} & \makecell{Predict \&\\Sim.\ Behav.} & \makecell{Strategic\\Decision-Making} & \makecell{Infer Subj.\\Traits} & \makecell{Apply\\Knowledge} & \multicolumn{2}{|c|}{\textbf{Eval.\ Level}} & \multirow{2}{*}{\makecell{\textbf{Public}\\\textbf{Benchmark}}} \\
\textbf{Targeted Variable} & \multicolumn{2}{c}{$y$} & $x$ & $\mathbb{K}$ & \makecell{Indiv.} & \makecell{Distrib.} & \\
\hline
\textbf{\bench} (Ours) & $\checkmark$ & $\checkmark$ & $\checkmark$ & $\checkmark$ & $\checkmark$ & $\checkmark$ & $\checkmark$ \\
\hline
SOUL-Index \citep{OdysSim} & $\checkmark$ & & & & $\checkmark$ & & $\checkmark$ \\
SocioBench \citep{sociobench} & $\checkmark$ & & & & $\checkmark$ & & $\checkmark$ \\
SubPOP \citep{subpop} & $\checkmark$ & & & & & $\checkmark$ & \\
SimBench \citep{simbench} & $\checkmark$ & & & & & $\checkmark$ & $\checkmark$ \\
GLEE \citep{glee} & & $\checkmark$ & & & $\checkmark$ & & $\checkmark$ \\
GTBench \citep{GTBENCH} & & $\checkmark$ & & & $\checkmark$ & & $\checkmark$ \\
GAMA-Bench \citep{GAMA} & & $\checkmark$ & & & $\checkmark$ & & $\checkmark$ \\
\hline
\end{tabular}}
\vspace{3pt}
\caption{Comparison of \bench with existing behavioral benchmarks. \bench is the only one evaluating all four capabilities at both individual and distributional levels.
}
\label{tab:benchmark-comparison}
\end{table*}

%% file: tables/main-results.tex
\begin{table*}[t]
\centering
\tiny
\setlength{\tabcolsep}{1.5pt}
\expandafter\let\csname endtabular*\endcsname\endtabular
\begin{tabular*}{\textwidth}{@{\extracolsep{\fill}}l c|cc|cc || cc|cc|cc || c}
\toprule
\textbf{Capability} & \multicolumn{11}{c||}{\textbf{Behavior Prediction \& Simulation}} & \multicolumn{1}{c}{\textbf{Strategic}} \\
\cmidrule(lr){2-12} \cmidrule(lr){13-13}
\textbf{Task} & \textbf{Game Behav. Sim.} & \multicolumn{2}{c|}{\textbf{Multi-Round Pred.}} & \multicolumn{2}{c||}{\textbf{Across-Ctx Pred.}} & \multicolumn{2}{c|}{\textbf{Demo. To Resp.}} & \multicolumn{2}{c|}{\textbf{Seq. Resp. Pred.}} & \multicolumn{2}{c||}{\textbf{Masked Resp. Pred.}} & \textbf{Game Play} \\
\cmidrule(lr){2-13}
\textbf{Metric} & \textbf{W $\downarrow$} & \textbf{MAE $\downarrow$} & \textbf{W $\downarrow$} & \textbf{MAE $\downarrow$} & \textbf{W $\downarrow$} & \textbf{Acc $\uparrow$} & \textbf{W $\downarrow$} & \textbf{Acc $\uparrow$} & \textbf{W $\downarrow$} & \textbf{Acc $\uparrow$} & \textbf{W $\downarrow$} & \textbf{Win Rate $\uparrow$} \\
\midrule
\textit{Lower bound for} $W$ & 2.0 & --- & 1.5 & --- & 1.1 & --- & 0.26 & --- & 0.25 & --- & 0.26 & --- \\
\midrule
Qwen3-4B & 26.9 & 21.7 & 13.1 & 27.9 & 20.1 & 24.3\% & 1.14 & 37.0\% & 0.65 & 37.2\% & 0.66 & 9\% \\
Llama3.3-70B & 24.2 & 27.0 & 15.3 & 26.4 & 14.9 & 28.4\% & 1.02 & 37.7\% & 0.59 & 43.9\% & 0.47 & 6\% \\
HY3 & 30.0 & 23.0 & 15.8 & 27.1 & 14.0 & 27.0\% & 0.80 & 42.1\% & 0.46 & 45.6\% & 0.41 & 10\% \\
DeepSeek V3.2 & \underline{16.9} & 20.9 & \underline{10.1} & 27.5 & \underline{12.3} & 27.2\% & 0.76 & 36.2\% & 0.54 & 40.6\% & 0.47 & 14\% \\
GLM-5.2 & 20.4 & 18.8 & \underline{10.1} & 25.7 & 18.7 & 27.0\% & 0.77 & 41.2\% & \underline{0.45} & 45.9\% & 0.38 & \underline{16\%} \\
\midrule
Claude Sonnet 4.6 & 24.6 & 18.7 & 10.6 & \underline{25.3} & 19.3 & 28.6\% & 0.86 & 41.8\% & \underline{0.45} & 46.7\% & \underline{0.36} & \textbf{17\%} \\
Claude Opus 4.6 & 26.0 & 19.3 & 11.7 & 25.9 & 20.1 & 29.7\% & 0.86 & 41.8\% & \textbf{0.43} & 45.1\% & 0.37 & 15\% \\
Claude Opus 5 & 26.7 & \textbf{17.7} & 11.5 & 25.8 & 15.4 & \underline{30.7\%} & 0.73 & 40.6\% & 0.48 & \textbf{49.0\%} & \textbf{0.34} & \underline{16\%} \\
GPT 4.1 & 27.5 & \underline{18.1} & \textbf{6.9} & 27.1 & 16.3 & 29.2\% & 0.79 & 39.0\% & 0.48 & 46.4\% & 0.38 & \textbf{17\%} \\
GPT 5.4 & 31.4 & 21.1 & 15.2 & 26.1 & 14.0 & 29.6\% & 0.90 & 42.4\% & 0.46 & 47.3\% & 0.38 & 8\% \\
GPT 5.6 Terra & 25.0 & 21.7 & 14.0 & 27.2 & 14.9 & \underline{30.7\%} & 0.90 & 41.0\% & 0.48 & 47.3\% & 0.40 & 11\% \\
GPT 5.6 Sol & 24.9 & 22.0 & 16.1 & 25.5 & 13.4 & \textbf{32.4\%} & 0.88 & \textbf{43.1\%} & 0.46 & 48.0\% & 0.40 & 10\% \\
Gemini 3.1 Flash & 31.3 & 25.5 & 18.6 & 29.4 & 17.3 & 30.5\% & 0.82 & 40.5\% & 0.52 & 46.3\% & 0.41 & 14\% \\
Gemini 3.1 Pro & 21.3 & 19.4 & 12.5 & 26.7 & \textbf{10.3} & 30.5\% & 0.75 & \underline{42.5\%} & 0.46 & \underline{48.5\%} & \underline{0.36} & 12\% \\
\midrule
Socrates-Llama3-8B-SFT & \textbf{16.3} & 39.5 & 21.3 & 37.7 & 14.6 & 24.0\% & \underline{0.55} & 23.5\% & 0.62 & 24.0\% & 0.67 & 2\% \\
Socrates-Llama3-8B-DPO & 26.4 & 27.3 & 19.7 & 27.7 & 21.4 & 24.7\% & 0.93 & 25.9\% & 0.96 & 28.4\% & 0.76 & 3\% \\
Socrates-Qwen-14B-SFT & 17.9 & 35.9 & 15.9 & 36.7 & 14.7 & 23.8\% & 0.69 & 23.8\% & 0.71 & 26.2\% & 0.69 & 2\% \\
Socrates-Qwen-14B-DPO & 20.0 & 20.5 & 14.4 & \textbf{24.8} & 17.7 & 23.9\% & 1.01 & 35.3\% & 0.71 & 37.8\% & 0.56 & 8\% \\
Centaur-70B & \underline{16.9} & 23.3 & 11.1 & 29.3 & 13.9 & 24.7\% & \textbf{0.43} & 33.1\% & 0.48 & 36.1\% & 0.49 & 4\% \\
OSim-8B & 21.9 & 22.7 & 12.4 & 29.4 & 21.2 & 29.4\% & 0.90 & 37.5\% & 0.63 & 37.8\% & 0.60 & 14\% \\
\bottomrule
\end{tabular*}

\vspace{6pt}

\begin{tabular*}{\textwidth}{@{\extracolsep{\fill}}l|cc|cc|cc||c|c}
\toprule
\textbf{Capability} & \multicolumn{6}{c||}{\textbf{Subject-trait Inference}} & \multicolumn{2}{c}{\textbf{Knowledge Application}} \\
\cmidrule(lr){2-9}
\textbf{Task} & \multicolumn{2}{c|}{\textbf{Demo. To Pers.}} & \multicolumn{2}{c|}{\textbf{Across-Dim Pers. Pred.}} & \multicolumn{2}{c||}{\textbf{Pers. To Demo.}} & \textbf{Sci. Workflow Pred.} & \textbf{Econ. Contest} \\
\cmidrule(lr){2-9}
\textbf{Metric} & \textbf{MAE$\downarrow$} & \textbf{W$\downarrow$} & \textbf{MAE$\downarrow$} & \textbf{W$\downarrow$} & \textbf{MAE$\downarrow$} & \textbf{W$\downarrow$} & \textbf{BLEURT$\uparrow$} & \textbf{Acc. $\uparrow$} \\
\midrule
\textit{Lower bound for} $W$ & --- & 0.65 & --- & 0.65 & --- & 0.42 & --- & --- \\
\midrule
Qwen3-4B & 7.3 & 5.14 & 7.6 & 5.21 & 11.2 & 10.44 & 0.450 & $73.9\%{\scriptstyle\,\pm\,1.3}$ \\
Llama3.3-70B & 6.9 & 5.18 & 8.3 & 5.19 & \underline{9.3} & 7.67 & 0.430 & $65.8\%{\scriptstyle\,\pm\,0.6}$ \\
HY3 & 7.1 & 4.91 & 7.7 & 5.32 & 13.9 & 7.21 & \underline{0.474} & $91.3\%{\scriptstyle\,\pm\,0.9}$ \\
DeepSeek V3.2 & 7.3 & 5.12 & 7.5 & 4.61 & 10.7 & 7.57 & 0.430 & $74.4\%{\scriptstyle\,\pm\,1.6}$ \\
GLM-5.2 & 7.0 & 5.47 & 7.4 & 4.51 & 10.5 & 4.27 & 0.458 & $92.7\%{\scriptstyle\,\pm\,1.4}$ \\
\midrule
Claude Sonnet 4.6 & 7.6 & 5.83 & 7.2 & 4.37 & 14.5 & 8.72 & 0.472 & $83.1\%{\scriptstyle\,\pm\,1.3}$ \\
Claude Opus 4.6 & 7.0 & 5.18 & \underline{6.7} & 4.49 & 13.3 & 8.47 & \textbf{0.477} & $\mathbf{95.6}\%{\scriptstyle\,\pm\,0.6}$ \\
Claude Opus 5 & \textbf{6.3} & 4.96 & \textbf{6.0} & \underline{3.65} & \textbf{9.1} & 3.70 & 0.440 & $\underline{95.2}\%{\scriptstyle\,\pm\,0.0}$ \\
GPT 4.1 & 6.9 & 4.92 & 7.2 & 3.93 & 10.6 & 4.01 & 0.460 & $78.4\%{\scriptstyle\,\pm\,0.8}$ \\
GPT 5.4 & 7.4 & 5.70 & 7.3 & 4.40 & 10.5 & 4.05 & 0.462 & $\mathbf{95.6}\%{\scriptstyle\,\pm\,0.6}$ \\
GPT 5.6 Terra & 7.5 & 6.73 & 7.2 & 4.59 & 10.0 & \underline{3.14} & 0.465 & $92.7\%{\scriptstyle\,\pm\,1.0}$ \\
GPT 5.6 Sol & 7.6 & 6.84 & 7.0 & 4.39 & 11.5 & 7.19 & 0.468 & $94.8\%{\scriptstyle\,\pm\,0.6}$ \\
Gemini 3.1 Flash & 7.1 & 4.78 & 7.1 & 4.98 & 14.0 & 8.17 & 0.432 & $77.6\%{\scriptstyle\,\pm\,1.7}$ \\
Gemini 3.1 Pro & \underline{6.4} & 4.38 & 6.8 & \textbf{3.60} & 10.5 & \textbf{1.67} & 0.470 & $\mathbf{95.6}\%{\scriptstyle\,\pm\,0.4}$ \\
\midrule
Socrates-Llama3-8B-SFT & 13.3 & 10.38 & 12.4 & 9.47 & 12.6 & 4.81 & 0.026 & $0.0\%{\scriptstyle\,\pm\,0.0}$ \\
Socrates-Llama3-8B-DPO & 8.7 & 6.14 & 8.6 & 6.73 & 9.6 & 5.97 & 0.426 & $39.7\%{\scriptstyle\,\pm\,4.6}$ \\
Socrates-Qwen-14B-SFT & 9.0 & \underline{3.77} & 9.8 & 4.90 & 11.0 & 3.69 & 0.024 & $0.0\%{\scriptstyle\,\pm\,0.0}$ \\
Socrates-Qwen-14B-DPO & 8.1 & 6.46 & 9.0 & 6.85 & 10.5 & 7.83 & 0.448 & $70.3\%{\scriptstyle\,\pm\,0.6}$ \\
Centaur-70B & 9.8 & \textbf{3.23} & 9.3 & 3.83 & 10.6 & 5.18 & 0.430 & $37.7\%{\scriptstyle\,\pm\,2.8}$ \\
OSim-8B & 7.0 & 5.69 & 7.5 & 5.13 & 10.9 & 9.04 & 0.449 & $68.5\%{\scriptstyle\,\pm\,1.4}$ \\
\bottomrule
\end{tabular*}
\vspace{3pt}
\caption{Benchmarking results on \bench.
\textbf{Bold} and \underline{underline} denote the best and second best performance.}
\label{tab:main-results}
\label{tab:econ-survey-summary}
\label{tab:infer-knowledge}
\end{table*}

%% file: sections/benchmarking/baselines.tex
We benchmark three types of foundation models representing different model types:
(1) open-source general-purpose LLMs, (2) proprietary general-purpose LLMs, and (3) behavior foundation models that are trained on behavior-related data.

\textbf{Open-source LLMs.}
We include open-source LLMs with various model sizes, including
Qwen3-4B~\citep{qwen3_report}, Llama-3.3-70B~\citep{llama3_report},
HY3~\citep{hy3} (295B), DeepSeek V3.2~\citep{deepseek-v3.2} (671B), and GLM-5.2~\citep{glm5} (753B), spanning from small dense models to bigger mixture-of-experts models.

\textbf{Proprietary LLMs.}
We include three families of widely used frontier proprietary models. Within each family we include different capability tiers. For Claude, we include Sonnet 4.6~\citep{claude-sonnet-4-6}, Opus 4.6~\citep{claude-opus-4-6} and Opus 5~\citep{claude-opus-5}. For GPT, we include leading reasoning models, GPT-5.4~\citep{gpt-5-4}, GPT 5.6 Terra and Sol~\citep{gpt-5-6}, together with GPT-4.1~\citep{gpt-4-1} as the strongest non-reasoning GPT model. For Gemini, we include Gemini 3.1 Flash and Gemini 3.1 Pro~\citep{gemini-3-1-pro}. We use the default inference hyperparameters for all models.

\textbf{Behavioral foundation models.}
This category includes foundation models fine-tuned on behavior-related data. In the \bench evaluation, we include three representative model families. Centaur-70B~\citep{centaur} is fine-tuned on a cognitive-psychology corpus, while Socrates~\citep{socsci210} is fine-tuned on social-science survey data. OSim-8B~\citep{OdysSim} is fine-tuned on diverse human interaction data, including conversation, social simulation, and role-play.

%% file: sections/behaviorbench/eval-setup.tex
Beyond heterogeneous per-task metrics, we aim to derive a unified score to compare models across tasks.
We treat each task as a pairwise comparison among the models, inspired by the Holistic Evaluation of Language Models (HELM) leaderboards~\citep{helm}. Concretely, we define a model's \textit{win rate} as the fraction of other models it outperforms on a given task (based on the task-specific metric), and \textit{mean win rate} as the average of win rates across tasks.
The leaderboards include the mean win rate rankings at both the individual and distributional levels.

%% file: sections/results/capabilities.tex
Table~\ref{tab:main-results} reports model performance across \bench tasks, grouped by capabilities. We summarize the main findings below. Appendix~\ref{app:predict-simulate-details} reports per-game results for the game tasks.

\begin{wraptable}{r}{0.5\textwidth}
    \centering
    \input{tables/diversity-offset}
    \caption{%
    Standard-deviation ratio and standardized bias of model predictions,
    averaged across the nine tasks with distributional evaluation metrics. Best in each column is \textbf{bold} (deviation ratio closest to $1$ and lowest bias), second
    best \underline{underlined}.
    }
    \label{tab:diversity-offset}
    \vspace{-5pt}
\end{wraptable}
\paragraph{\bench remains challenging for frontier models.}
At the individual level, the best accuracies on Demo.\ To Resp., Seq.\ Resp.\ Pred., and Masked Resp.\ Pred.\ are only $32.4\%$, $43.1\%$, and $49.0\%$ respectively. On Multi-Round Pred. and Across-Ctx Pred. tasks, even the best performance is off by $17.7$ and $24.8$ in MAE, when the range is $0$--$100$. For strategic game play, the best models only win $17\%$ of the Beauty Contest sessions they play. On distributional metrics, we follow previous work \cite{subpop,xie2025distributional} and compute a lower bound on $W$ by bootstrapping the human responses (first row of Table~\ref{tab:main-results}). We find that the best $W$ on each task is $4.6$ times larger than this lower bound on average, indicating there is still substantial room for improvement.

\paragraph{Gemini 3.1 Pro and Claude Opus 5 lead \bench.}
As shown in Figure~\ref{fig:ind-vs-dist}, Claude Opus 5 ranks 1st of the 20 models at the individual level and 3rd at the distributional level, and Gemini 3.1 Pro ranks 2nd and 1st. They are the only two models in the top three at both levels.

\paragraph{Frontier proprietary models retain an advantage on contest-style economic reasoning.}
Claude Opus 4.6, GPT 5.4, and Gemini 3.1 Pro each achieve $95.6\%$ accuracy on the economics contest task. In contrast, Centaur-70B and most Socrates variants perform notably worse on knowledge-intensive tasks, especially the economics contest, suggesting that models fine-tuned for behavioral prediction do not automatically acquire broader behavioral-science reasoning ability.

\subsection{Individual-level and distributional performance do not always align} \label{sec:results-indiv-vs-dist}

Model performance at the individual level does not necessarily guarantee a strong distributional performance.
For example, Claude Opus 4.6 and GPT 5.6 Sol rank 3rd and 4th at the individual level, but fall to 9th and 11th at the distributional level. By contrast, Centaur-70B ranks 2nd at the distributional level, but only ranks 18th at the individual level.
To examine how behavioral foundation models better preserve the diversity of human responses than general-purpose models, we report two statistics for models that rank among the top five on either the individual-level or distributional leaderboard:
\begin{itemize}[leftmargin=*,itemsep=0pt,topsep=2pt]
\item $\sigma_{\text{model}}/\sigma_{\text{human}}$, which compares the variation in model predictions with that in human responses. A value closer to 1 indicates that the model better captures the diversity of human responses.
\item  $\frac{|\mu_{\text{model}}-\mu_{\text{human}}|}{\sigma_{\text{human}}}$, which measures the difference between the model and human population means, normalized by the standard deviation of human responses. A lower value indicates better alignment with the human population mean.
\end{itemize}

We average both statistics across the nine distributional metrics in \bench. As shown in Table~\ref{tab:diversity-offset}, \textbf{general-purpose LLMs tend to underestimate the diversity of human responses, even though the leading ones predict the population mean more accurately. Behavioral foundation models, by contrast, better preserve response diversity}.
The general-purpose models in Table~\ref{tab:diversity-offset} reproduce only $0.53$--$0.70$ of the human standard deviation, against $0.93$ for Centaur-70B.

%% file: tables/diversity-offset.tex
\begingroup
\scriptsize
\setlength{\tabcolsep}{5pt}
\renewcommand{\arraystretch}{1.25}
\begin{tabular}{lcc}
\toprule
 & $\sigma_{\text{model}}/\sigma_{\text{human}}$ & $\frac{|\mu_{\text{model}}-\mu_{\text{human}}|}{\sigma_{\text{human}}}$ $\downarrow$\\
\midrule
Gemini 3.1 Pro & $\underline{0.70}$ & $\underline{0.31}$ \\
Centaur-70B & $\textbf{0.93}$ & $0.38$ \\
Claude Opus 5 & $0.53$ & $\textbf{0.26}$ \\
GPT 4.1 & $0.62$ & $0.38$ \\
GLM-5.2 & $0.61$ & $0.34$ \\
Claude Opus 4.6 & $0.54$ & $0.42$ \\
GPT 5.6 Sol & $0.63$ & $0.46$ \\
\bottomrule
\end{tabular}
\endgroup

%% file: sections/discussion/generalization-summary.tex
\begin{wrapfigure}{r}{0.5\textwidth}
    \centering
    \includegraphics[width=\linewidth]{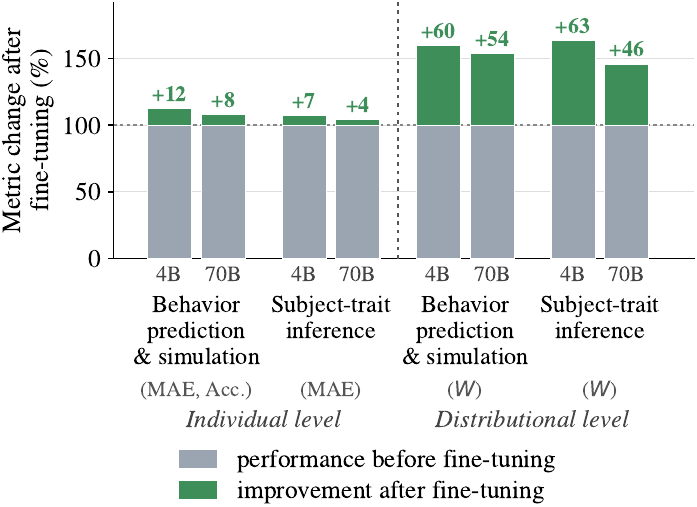}
    \caption{%
    Performance improvement after fine-tuning.
    Both Qwen3-4B and Llama-3.3-70B improve more at the distributional level than the individual level.}
    \label{fig:generalization-three-tier}
    \vspace{-10pt}
\end{wrapfigure}

In Section~\ref{sec:results-indiv-vs-dist}, we find that general-purpose LLMs and behavior foundation models exhibit uneven performance at distributional and individual levels. In this section, we ask the question: is it possible to optimize performance on these two evaluation levels at the same time? To answer this question, we fine-tune general-purpose LLMs on diverse behavior data targeting the capabilities in \bench.

\subsection{Training Data and Setup}

We fine-tune two widely used open-weight backbone models, Qwen3-4B~\citep{qwen3_report} and Llama-3.3-70B~\citep{llama3_report}, on diverse behavioral data. Following the task formulation in Section~\ref{sec:evaluation-tasks}, we instantiate fine-tuning tasks that target the capabilities evaluated by \bench.

Compared with the fine-tuning datasets of previous works on behavior foundation models, our curated dataset prioritizes breadth of data sources over depth within each source. Specifically, we include all three data sources described in Section~\ref{sec:behavioral-data}.
In contrast, Centaur-70B~\citep{centaur} uses only experimental data, while Socrates~\citep{socsci210} uses only survey data.
In particular, our fine-tuning dataset includes 649,394 data points across 11 fine-tuning tasks. We note that the training data is disjoint from \bench.
Details on the backbone models, the fine-tuning schema, and the fine-tuning data are described in Appendix~\ref{app:models-befm}.

\subsection{Improvement on Distributional Level Is Larger than Individual Level}

As shown in Figure~\ref{fig:generalization-three-tier}, both fine-tuned models improve over their respective backbone models.
The gains are bigger at the distributional level ($46\%$--$63\%$), while individual-level prediction improves consistently but by much smaller margins ($4.0\%$--$12.3\%$).
One explanation is that the two levels involve different prediction targets. Distributional metrics compare only the predicted behavior distribution with the human distribution, $p(y \mid c)$, whereas individual-level metrics require accurate subject-specific predictions, $p(y \mid x, c)$. Our experiment shows that fine-tuning on human behavior data may improve predictions of the population distribution more than predictions for individual subjects.
Overall, the strong performance of fine-tuned models implies that individual-level accuracy and faithful distributional alignment can be improved simultaneously, while also revealing substantial room for improvement.

%% file: sections/conclusion.tex
We introduce \bench, a comprehensive benchmark that evaluates foundation models across four capabilities with metrics at both individual and distributional levels. Our evaluation reveals that current foundation models exhibit uneven strengths across tasks.
Particularly, general-purpose frontier proprietary LLMs excel at individual-level prediction and knowledge-intensive tasks. In contrast, behavioral foundation models perform strongly on distributional alignment.
Our fine-tuning experiment suggests that models can achieve both individual-level accuracy and faithful distributional alignment, while showing that foundation models still have substantial room to improve for behavioral science.
More broadly, by emphasizing distributional alignment, \bench can help assess whether AI systems reflect the heterogeneity of human populations rather than overfitting to majority or average behaviors.
Our results establish \bench as a foundation for developing and assessing behaviorally aligned AI systems.

%% file: sections/related/foundation-models.tex
General foundation models, including GPT and Llama, are increasingly being applied to behavioral science domains such as psychology, sociology, and economics~\citep{demszky2023using,meng2024ai,bail2024can}. These models have shown promise in predicting survey responses~\citep{argyle2023out,grossmann2023ai}, enabling automated content analysis~\citep{wu2023large,yang2023large,gilardi2023chatgpt}, supporting agent-based modeling~\citep{park2023generative,park2024generative}, and inferring experimental results~\citep{hewitt2024predicting,lippert2024can}. However, recent work has also revealed important limitations in directly applying general-purpose LLMs to behavioral science tasks. For example, these models often fail to capture population variance~\citep{bisbee2024synthetic}, exhibit sensitivity or bias toward specific data formats~\citep{dentella2023systematic}, and underrepresent certain populations~\citep{opinionqa}. This mismatch between the needs of behavioral science and the design of general foundation models calls for domain-specific models that better align with behavioral science principles, data, and evaluation goals.

A small but growing body of work has begun to address this need. Monad~\citep{rychalska2023synerise}, for example, learns dense user representations from streams of clicks, transactions, and other behavioral events. Centaur~\citep{centaur} models human responses in cognitive-psychology experiments, including tasks involving decision-making, memory, supervised learning, and Markov decision processes. \modelfamily models~\citep{BeFM-1}, trained on experimental, survey, and literature data, demonstrate capabilities in predicting human behavior, inferring subject characteristics, generating insights about contextual factors, and applying behavioral science knowledge to problem-solving. Socrates~\citep{socsci210} adapts language models to social-science survey data.
However, despite this growing body of work, we still lack a systematic understanding of how well foundation models perform across diverse behavioral science tasks, contexts, and populations.

%% file: sections/related/ai-behavioral-science.tex
AI behavioral science is inherently bidirectional---not only can AI serve as a tool in behavioral science research as discussed above, but behavioral science methods can also be used to study the behaviors, biases, and emergent properties of AI systems themselves~\citep{jackson2025AI}.
For example, \citet{aher2023using,mei2024turing} and \cite{xie2024different} have conducted variations on Turing tests, comparing LLMs' decisions in economic games to human data. Similarly, \citet{horton2023large, lore2024strategic} investigated the strategic behavior of LLMs, while \citet{chen2023emergence} assessed GPT's economic rationality. Other studies applied behavioral frameworks such as theory of mind to LLMs \citep{strachan2024testing}, or examined biases in AI-generated survey responses \citep{salecha2024large}.

While these two directions---using AI for behavioral science and applying behavioral science to AI---have each advanced, a unifying approach remains missing. Specifically, there is a lack of models that function both as tools for understanding human behavior and as platforms for behavioral interrogation of AI systems themselves.

%% file: sections/related/benchmarks.tex
A growing line of work has developed benchmarks for evaluating how well language models can reproduce human responses in surveys and social science experiments. OpinionQA \citep{opinionqa} compares model outputs against U.S. public opinion surveys to study whose opinions language models reflect across demographic groups. SimBench \citep{simbench} harmonizes 20 datasets spanning moral dilemmas, economic choices, and psychological assessments into a single standardized benchmark, and measures how closely model predictions match group-level human response distributions. SubPOP \citep{subpop} targets the prediction of public opinion distributions, showing that fine-tuning on survey data substantially improves alignment with held-out human responses. SocioBench \citep{sociobench} models human behavior in sociological surveys, while SocSci210 \citep{socsci210} fine-tunes language models on social science experiments to predict participant-level and distributional-level outcomes. These benchmarks have established survey response prediction as a central task for evaluating LLMs as proxies for human respondents. However, most prior work focuses primarily on this single task formulation without exploring alternative formulations such as predicting missing survey responses.

Beyond predicting responses in surveys and social science experiments, recent works also evaluate LLMs in economic and strategic settings. GTBench \citep{GTBENCH} evaluates the strategic reasoning limitations of LLMs across a suite of classic game-theoretic scenarios, and GAMA-Bench \citep{GAMA} embeds competing language models within multi-agent gaming environments. CHBench \citep{chbench} grounds evaluation in cognitive hierarchy theory, while Economics Arena \citep{econarena} benchmarks LLMs across a range of economic decision tasks. GLEE \citep{glee} provides a unified framework and benchmark for language-based economic environments such as bargaining, negotiation, and persuasion games. Most of these benchmarks have LLMs play against other LLMs or against heuristics, with no human opponents.
In our work, we use human play records from economic games to construct a proxy evaluation for LLMs playing against human players.

%% file: sections/appendix/data-details.tex
\section{Data Details}
\label{app:data-details}

\subsection{Data Sources}
\label{app:data-sources}

This subsection provides a more detailed walkthrough of the data sources introduced in Section~\ref{sec:behavioral-data}.

\paragraph{Experimental data.}
Experimental data document the behavioral patterns of varied participant groups under controlled conditions, providing crucial insights into how individuals and groups make decisions and engage socially.
We utilize a collection of experiments from MobLab, a platform for behavioral science and economic games. After cleaning, this source contains 379{,}754 behavioral observations from 76{,}035 unique subjects collected over a nine-year period (2015 to 2023)~\citep{mei2024turing}. The collection covers seven classic economic games across nine scenarios: Dictator, Ultimatum (Proposer and Responder), Trust (Investor and Banker), Public Goods, Bomb Risk, Beauty Contest, and Push/Pull (Prisoner's Dilemma).

\paragraph{Survey data.}
Survey data offer self-reported information about personality traits, attitudes, and preferences. They primarily characterize subject attributes ($x$ in Equation~\ref{eqn:framework}), though they may also contain self-reported behaviors $\tilde{y}$, which serve as a proxy for observed behaviors $y$ and can be leveraged to simulate responses to surveys not yet conducted. Such responses may be elicited either without any specified context or under particular pre-defined scenarios $c$ embedded in the survey items. We adopt the publicly available Big Five Personality Test dataset\footnote{\raggedright\url{https://www.kaggle.com/datasets/lucasgreenwell/ocean-five-factor-personality-test-responses}.}, which contains demographic attributes (race, age, and gender) of participants along with their single-choice responses to 50 items spanning five personality dimensions: openness, conscientiousness, extraversion, agreeableness, and neuroticism. The dataset comprises 19,630 subjects and a total of 981,500 survey responses.

\paragraph{Literature data.}
The behavioral science literature captures established theories and empirical findings on human behavior, serving as a proxy for the behavioral knowledge $\mathbb{K}$ in Equation~\ref{eqn:framework}. We gather titles and abstracts from two leading interdisciplinary venues for behavioral research: the \emph{American Economic Review} (AER) and \emph{Nature Human Behaviour} (NHB). The resulting corpus consists of 6{,}515 AER articles spanning 1911 to 2025 and 871 NHB articles spanning 2016 to 2025. Following the approach of MASSW~\citep{massw}, we summarize each title--abstract pair into a structured research workflow capturing the research context, key idea, method, outcome, and anticipated impact using GPT-5~\citep{gpt-5}.

\paragraph{Economics contest data.}
We also use multiple-choice questions from the International Economics Olympiad (IEO) as an evaluation-only source for economics problem solving. This source differs from the experimental and survey datasets because it does not record human behavioral variation. Instead, it evaluates whether models can represent, reason about, and apply domain-specific economic knowledge to unseen problems.

\subsection{Data Quality Control}
\label{app:data-quality}

Each data source passes a cleaning and validation pipeline before any task instance is constructed. We report the filters and the retention rate for each source below.

\paragraph{MobLab economic games.} Raw move records pass a four-level filter: (L1) records carrying an invalid placeholder value in the move or round result, together with all subsequent rounds of the same subject; (L2) moves outside the action range implied by the game instruction; (L3) sessions with a non-standard game configuration (endowment, group size, or role), excluded so that the payoff structure is comparable across subjects; and (L4) rounds that create a gap in a subject's round sequence after L1--L3. Of the $569{,}795$ raw records, $379{,}754$ ($66.6\%$) remain, covering $76{,}035$ of $85{,}343$ subjects.

\paragraph{Big Five survey.} We remove one respondent who left all $50$ items blank and $88$ respondents reporting an implausible age ($\geq 80$), retaining $19{,}630$ of $19{,}719$ respondents ($99.5\%$). The retained respondents span $158$ countries with a median age of $22$.

\paragraph{Literature corpus.} Of $9{,}255$ AER records retrieved from OpenAlex, $6{,}515$ have both a title and an abstract and are retained; all $871$ NHB records have both. The GPT-5 workflow extraction returns a set of five aspects for $6{,}514/6{,}515$ AER and $870/871$ NHB papers. After this filter, the $257$ papers published in 2025 yield $1{,}084$ of $1{,}285$ possible evaluation instances.

\paragraph{Economics contest.} The source contains $130$ multiple-choice problems across seven IEO years (2018, 2019, and 2021--2025). We remove the $6$ problems whose answer depends on a figure that cannot be reproduced in text, leaving the $124$ problems used for evaluation.

\subsection{Data Splits}
\label{app:data-split}

This subsection describes how each raw source from Section~\ref{app:data-sources} is partitioned into the fine-tuning set and the \bench evaluation set, and how we sample task instances from each partition. Table~\ref{tab:data-split-summary} summarizes the per-task entry counts and the split rule used. Across all sources, the partition is constructed so that no subject (or paper) appears on both the fine-tuning and evaluation side of any task.

\paragraph{MobLab economic games.} We split unique participants (identified by a unique \texttt{UserID}) 70/10/20 into train, validation, and test, so a participant's complete history of game play remains within a single partition. The Game Behavior Sim., Multi-Round Pred., Across-Ctx Pred., and Strategic Game Play tasks are all generated from these splits. Raw game logs are first cleaned by a filtering pipeline that removes invalid game choices.

\paragraph{The Push/Pull game (held-out).} The Push/Pull game is excluded from fine-tuning entirely. When constructing the Push/Pull evaluation, we additionally drop any subject that appears in the MobLab train or validation split, so no participant the model has seen during fine-tuning contributes to this held-out evaluation.

\paragraph{Big Five survey.} We use a 90/10 split over subjects. All Big Five tasks (Demo.\ to Resp., Seq.\ Resp.\ Pred., Masked Resp.\ Pred., Demo.\ to Pers., Across-Dim Pers.\ Pred., and Pers.\ to Demo.) draw their fine-tuning and evaluation instances from this single participant-level split.

\paragraph{Scientific workflows.} We split AER and NHB papers by publication year. Papers published in 2025 form the workflow-prediction test set, and all earlier years form the fine-tuning set. This temporal cut reflects the practical use case of generalizing to newly published papers.

\paragraph{Economics contest (IEO).} The IEO multiple-choice problems serve as an evaluation-only source and have no corresponding fine-tuning data.

\paragraph{Sampling and downsampling.} For tasks where the natural enumeration produces far more instances than we need for either fine-tuning or evaluation, we downsample to a fixed target size. All Big Five evaluation tasks are capped at 1K examples. MobLab game tasks use random sampling per game, with a cap of 200, 500, and 750 examples per game for Game Behav.\ Sim., Multi-Round Pred., and Across-Ctx Pred.\ respectively. The Strategic Game Play evaluation set is capped at 1K examples. The final per-task counts after sampling are reported in Table~\ref{tab:data-split-summary}.

\input{tables/data-split-summary}

\subsection{Evaluation Tasks in \bench}
\label{app:evaluation-data}

For each evaluation task in \bench, we briefly describe the task below and provide its prompt template.

\paragraph{Single-round behavior simulation (Game Behav.\ Sim.).}
The model is shown the rules of a single MobLab economic game and asked for its first-round move. The game-specific instructions cover the Dictator game, the Ultimatum game (as Proposer or Responder), the Trust game (as Investor or Banker), the Public Goods game, the Bomb Risk game, the Beauty Contest game, and the Push/Pull game (Prisoner's Dilemma) (\hyperref[prompt:games-firstround]{[PROMPT]}).

\paragraph{Multi-round behavior prediction (Multi-Round Pred.).}
The model is shown the game instruction together with the subject's prior rounds in the same session, and predicts the next round's move (\hyperref[prompt:games-multiround]{[PROMPT]}).

\paragraph{Across-context single-round prediction (Across-Ctx Pred.).}
The model is shown a subject's first-round moves in several other games and asked to predict the subject's first-round move in a target game (\hyperref[prompt:games-acrossctx]{[PROMPT]}).

\paragraph{Survey response prediction given demographics (Demo.\ to Resp.).}
The model is given a subject's demographics and asked to predict one Big Five item of this subject on a five-point Likert scale (\hyperref[prompt:bigfive-singleresp]{[PROMPT]}).

\paragraph{Sequential survey response prediction (Seq.\ Resp.\ Pred.).}
The model is shown the subject's previous answers in the original survey order (interleaved across the five dimensions) and predicts the subject's response to the next item (\hyperref[prompt:bigfive-seq]{[PROMPT]}).

\paragraph{Masked survey response prediction (Masked Resp.\ Pred.).}
The model is shown 49 of a subject's 50 Big Five item responses, grouped by dimension, and predicts the held-out item (\hyperref[prompt:bigfive-acrossitem]{[PROMPT]}).

\paragraph{Personality score from demographics (Demo.\ to Pers.).}
The model is given a subject's demographics and predicts the subject's score on one Big Five dimension (\hyperref[prompt:bigfive-dim]{[PROMPT]}).

\paragraph{Across-dimension personality prediction (Across-Dim Pers.\ Pred.).}
The model is given a subject's scores on four Big Five dimensions and predicts the score on the held-out dimension (\hyperref[prompt:bigfive-leaveone]{[PROMPT]}).

\paragraph{Age prediction from personality (Pers.\ to Demo.).}
The model is given a subject's scores on all five Big Five dimensions and predicts the subject's age (\hyperref[prompt:bigfive-age]{[PROMPT]}).

\paragraph{Strategic game play.}
The model plays the Beauty Contest game against human players. Given the play history and the group average from each prior round, the model picks an integer in $[0, 100]$ that aims to be closest to two-thirds of the next group average (\hyperref[prompt:beauty-contest]{[PROMPT]}).

\paragraph{Scientific workflow prediction (Sci.\ Workflow Pred.).}
Five generation sub-tasks (idea generation, method recommendation, outcome prediction, impact prediction, and title prediction), each predicting one workflow aspect from the preceding aspects of the same paper (\hyperref[prompt:sciworkflow]{[PROMPT]}).

\paragraph{Economics contest (IEO).}
The model selects A/B/C/D for each multiple-choice problem from the International Economics Olympiad (\hyperref[prompt:ieo]{[PROMPT]}).

%% file: tables/data-split-summary.tex
\begin{table*}[t]
\centering
\small
\setlength{\tabcolsep}{4pt}
\def\arraystretch{1.2}
\resizebox{\textwidth}{!}{%
\begin{tabular}{p{4.2cm}p{5.5cm}rrp{2.5cm}}
\toprule
\textbf{Capability} & \textbf{Task} & \textbf{\# Train} & \textbf{\# Eval} & \textbf{Train/Test Split} \\
\midrule

\multirow{6}{4.2cm}{\textbf{Behavior prediction and simulation}}
& Game Behav.\ Sim. & 78.8K & 1,800 & by subject \\
& Multi-Round Pred. & 57.9K & 3,623 & by subject \\
& Across-Ctx Pred. & 44.1K & 6,262 & by subject \\
& Demo.\ To Resp. & 88.3K & 1,000 & by subject \\
& Seq.\ Resp.\ Pred. & 36.0K & 1,000 & by subject \\
& Masked Resp.\ Pred. & 40.0K & 1,000 & by subject \\
\midrule

\multirow{3}{4.2cm}{\textbf{Subject-trait inference}}
& Demo.\ To Pers. & 88.3K & 1,000 & by subject \\
& Across-Dim Pers.\ Pred. & 88.3K & 1,000 & by subject \\
& Pers.\ To Demo. & 17.7K & 1,000 & by subject \\
\midrule

\textbf{Strategic decision-making}
& Strategic Game Play & 85.2K & 1,000 & by subject \\
\midrule

\multirow{2}{4.2cm}{\textbf{Behavioral knowledge application}}
& Sci.\ Workflow Pred. & 24.7K & 1,084 & by publication year \\
& Economics Contest & --- & 124 & eval only \\
\midrule
\multicolumn{2}{l}{\textbf{Total}} & \textbf{649.4K} & \textbf{19,893} & \\
\bottomrule
\end{tabular}}
\vspace{5pt}
\caption{Training and evaluation data summary.}
\label{tab:data-split-summary}
\end{table*}

%% file: sections/appendix/metrics-details.tex
\section{Metrics Details}
\label{app:metrics-details}

In tasks constructed from economic games, players' choice ranges differ across games (e.g., in the Dictator game, a player can choose a number between 0 and 100, but for the Public Goods game, a player can choose between 0 and 20). We normalize the players' choices into range $[0, 100]$. We then average these normalized $W$ across all games.
A lower $W$ score indicates that the predicted distributions are closer to observed human behavior distributions.

\paragraph{Lower bound for $W$.}

Following \citet{subpop}, we estimate the lower bound Wasserstein distance by bootstrapping the human respondents.
For each task, we sample (with replacement) as many human responses as observed, and compute the Wasserstein distance between the observed human distribution and the bootstrapped distribution, averaging over evaluation units (a game, a survey item, or a personality dimension) in the same way as the model scores. We repeat this $1000$ times and report the mean as the \emph{lower bound for $W$} row of Table~\ref{tab:main-results}.

%% file: sections/befm-model.tex
\input{sections/befm-model/motivation}

\subsection{Backbone Models and Fine-tuning Schema}
\label{sec:models-befm-backbone}
\input{sections/befm-model/backbone}

\subsection{Fine-tuning Tasks and Data}
\label{sec:models-befm-training-data}
\input{sections/befm-model/training-data}

\subsection{Fine-tuning Data Recipe}
\label{app:fine-tuning-data}
\input{sections/befm-model/fine-tuning-data-recipe}

%% file: sections/befm-model/motivation.tex
\bench sets a higher bar for behavior foundation models, covering four capability categories and evaluating performance at both the individual and population levels. To test whether this bar can be met by behavioral fine-tuning rather than by scale alone, we fine-tune two open-weight LLMs on a broader set of behavioral tasks than prior behavioral fine-tuning efforts~\citep{BeFM-1}, with explicit coverage of diverse capabilities, behavioral contexts, and human populations.

%% file: sections/befm-model/backbone.tex
For fine-tuning, we use two widely-used open-weight backbones at different scales: Qwen3-4B~\citep{qwen3_report} and Llama-3.3-70B~\citep{llama3_report}. Using two backbones an order of magnitude apart in size lets us separate the effect of behavioral fine-tuning from the effect of model capacity, and each fine-tuned model is compared directly against its own untuned backbone. We use parameter-efficient supervised fine-tuning, applying Low-Rank Adaptation (LoRA)~\citep{hu2022lora} (rank 8, $\alpha=32$) to all linear layers via the \texttt{ms-swift} framework~\citep{ms-swift}. Both backbones are trained for one epoch on 4$\times$A100 80GB GPUs with a peak learning rate of $10^{-4}$, sequence length 2048, and effective batch size 128. Fine-tuning the 4B and 70B backbones takes around 60 and 600 A100 hours respectively.

%% file: sections/befm-model/training-data.tex
Inspired by \bench, we formulate fine-tuning tasks as described in Section~\ref{sec:evaluation-tasks} and curate fine-tuning data from the three sources described in Section~\ref{sec:behavioral-data}. Although fine-tuning shares the same task formulation with \bench, the fine-tuning data do not overlap with the \bench evaluation data. The splitting procedures are summarized in Appendix~\ref{app:data-split}.

Specifically, we instantiate 11 fine-tuning tasks organized by the general capabilities they target: behavior prediction and simulation, strategic decision-making, subject-trait inference, and the application of behavioral knowledge. The task mixture is deliberately broad. It exposes both backbones to all components in the framework of Equation~\ref{eqn:framework} (i.e., behavior $y$, subject traits $x$, context $c$, and behavioral science knowledge $\mathbb{K}$). This breadth is intended to encourage generalization across tasks, contexts, and populations, rather than specialization in any single task or dataset. Details on fine-tuning data are described in Section~\ref{app:fine-tuning-data}.

%% file: sections/befm-model/fine-tuning-data-recipe.tex
We instantiate eleven fine-tuning tasks that span the three data sources of Section~\ref{app:data-sources} and cover the four capability axes of \bench. All instances are drawn from the fine-tuning partitions defined in Section~\ref{app:data-split}, so no subject or paper that appears in the \bench evaluation set ever appears in fine-tuning.
Per-task counts are listed in Table~\ref{tab:data-split-summary}. We describe the tasks below grouped by the \bench capability they target.

\paragraph{Behavior prediction and simulation.} Six tasks. Three on MobLab games: single-round behavior simulation, multi-round behavior prediction, and across-context single-round prediction. Three on the Big Five survey: response prediction from demographics, sequential response prediction in survey order, and masked response prediction over the full 50-item survey.

\paragraph{Subject-trait inference.} Three Big Five tasks: dimension-score prediction from demographics, leave-one-dimension prediction across the five dimensions, and age prediction from the five dimension scores. The dimension-score-from-demographics task is upsampled by a factor of three in the final mix (three identical copies are shuffled together with the rest). This is because preliminary experiments show that this task requires more training to improve on the validation set.

\paragraph{Strategic decision-making.} One task: strategic Beauty Contest play. The Beauty Contest fine-tuning data uses a multi-winner expansion: when several integers tie as winners against the recorded other players in a round, each winning integer becomes a separate fine-tuning example (capped at 10 per round).

\paragraph{Behavioral knowledge application.} One task with five sub-tasks (idea generation, method recommendation, outcome prediction, impact prediction, and title prediction), each predicting one workflow aspect from the preceding aspects.

%% file: sections/appendix/direct-elicitation.tex
\section{Verbalized vs.\ Simulated Distributions}
\label{app:direct-elicitation}

The distributional metrics in Table~\ref{tab:main-results} score \emph{simulated distributions}: a model is repeatedly prompted for individual information, and the aggregated response distribution is compared against the human response distribution. An alternative is to elicit a \emph{verbalized distribution}: the model is asked a single question about the population and returns the full probability distribution as text, without any sampling~\citep{meister2025benchmarking}. We evaluate the verbalized setting on the nine distributional tasks of \bench.

\begin{figure*}[!t]
    \centering
    \includegraphics[width=\textwidth]{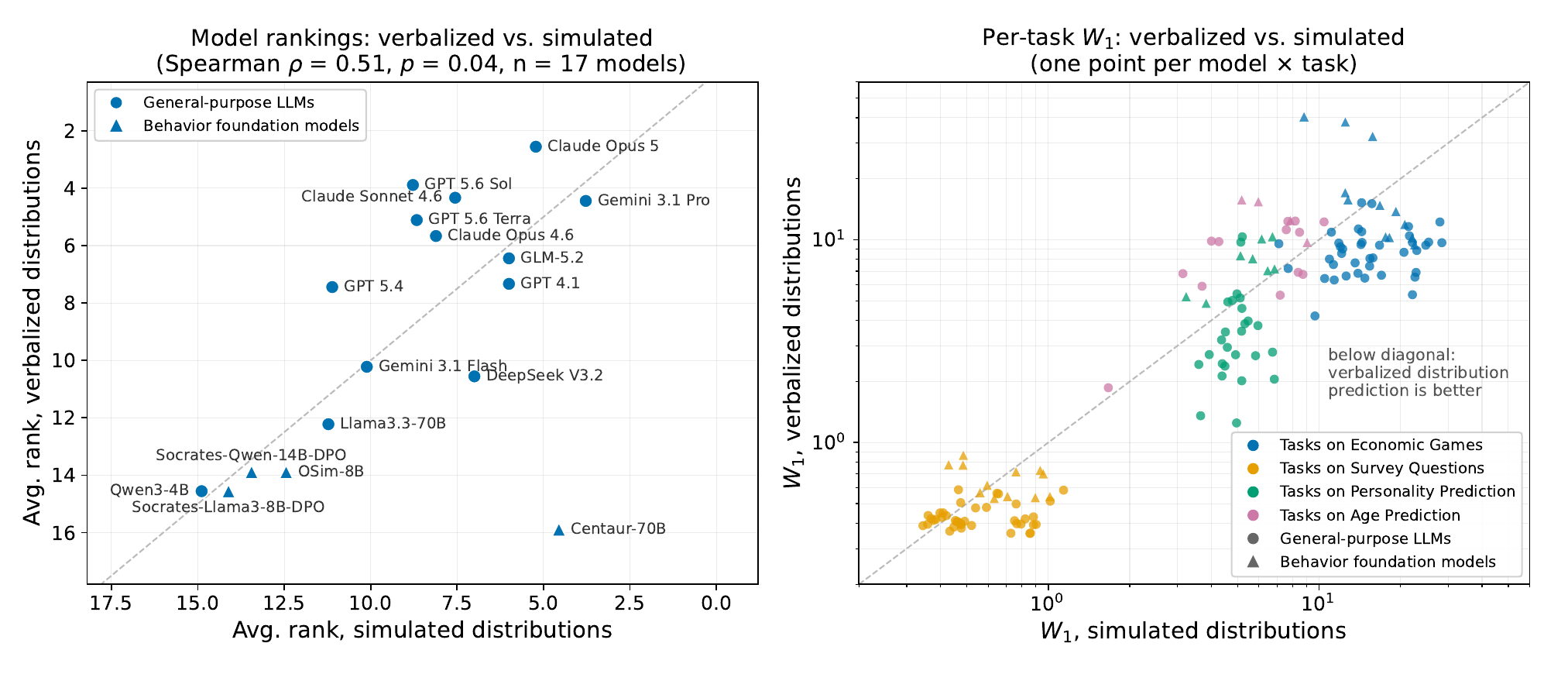}
    \caption{%
    Verbalized vs.\ simulated distributions.
    \textbf{(Left)} Average model rank for verbalized and simulated distributions.
    \textbf{(Right)} Per-task Wasserstein distance under the two settings. One
    point denotes the performance of one model in one task, colored by task family. Points below the
    diagonal are cases where the verbalized distribution is more accurate than
    the simulated one.}
    \label{fig:direct-vs-simulated}
\end{figure*}

Table~\ref{tab:direct-elicitation} reports the per-task $W$ of the verbalized distributions. We have two main observations:
\begin{itemize}[leftmargin=*,itemsep=0pt,topsep=2pt]
    \item \textbf{The model rankings under the two settings are positively correlated.} Average ranks over the nine tasks correlate at Spearman $\rho = 0.51$ ($p = 0.04$, Figure~\ref{fig:direct-vs-simulated} Left). This means that the two settings largely measure a shared capability: models that simulate the population distribution well also tend to describe it well. The behavior foundation models are the main exception, e.g., Centaur-70B ranks second-best on simulated distributions but last on verbalized ones.
    \item \textbf{Verbalized distributions tend to be more accurate than simulated ones.} Most general-purpose models attain a lower mean Wasserstein distance when verbalizing the distribution than when being sampled (Figure~\ref{fig:direct-vs-simulated} Right). This aligns with the knowledge-to-simulation gap reported for opinion surveys by \citet{meister2025benchmarking}: a model can often describe a population distribution more faithfully than it can sample from it. On the contrary, behavior foundation models tend to perform better with simulated distributions than verbalized distributions, which is different from general-purpose LLMs.
\end{itemize}

\input{tables/results-direct-elicitation}

%% file: tables/results-direct-elicitation.tex
\begin{table*}[!t]
\centering
\resizebox{\textwidth}{!}{%
\small
\begin{tabular}{lccccccccc|c}
\toprule
\textbf{Model} & \textbf{Game Behav.} & \textbf{Multi-Round} & \textbf{Across-Ctx} & \textbf{Demo.\ To Resp.} & \textbf{Seq.\ Resp.} & \textbf{Masked Resp.} & \textbf{Demo.\ To Pers.} & \textbf{Across-Dim Pers.} & \textbf{Pers.\ To Demo.} & \textbf{Avg.\ Rank} \\
\midrule
Qwen3-4B & 12.24 & 15.23 & 11.62 & 0.58 & 0.56 & 0.56 & 9.73 & 10.34 & 12.25 & 14.56 \\
Llama3.3-70B & 10.45 & 15.07 & 10.96 & 0.51 & 0.48 & 0.59 & 3.54 & 4.59 & 12.32 & 12.22 \\
DeepSeek V3.2 & 9.40 & 10.90 & 9.02 & 0.50 & 0.48 & 0.51 & 5.16 & 4.94 & 11.21 & 10.56 \\
GLM-5.2 & 6.68 & 8.03 & 6.47 & 0.40 & 0.41 & 0.42 & 3.98 & 3.51 & 9.79 & 6.44 \\
\midrule
Claude Sonnet 4.6 & 8.67 & 9.24 & 7.41 & \textbf{0.36} & 0.38 & \underline{0.40} & 2.68 & 3.21 & 6.75 & 4.33 \\
Claude Opus 4.6 & 9.69 & 9.63 & 8.14 & 0.36 & \textbf{0.37} & 0.42 & \underline{2.02} & \underline{2.38} & 10.89 & 5.67 \\
Claude Opus 5 & 8.85 & \textbf{6.34} & 7.69 & \underline{0.36} & \underline{0.38} & \textbf{0.39} & \textbf{1.25} & \textbf{1.36} & 5.90 & \textbf{2.56} \\
GPT 4.1 & 9.73 & 9.56 & 9.70 & 0.40 & 0.40 & 0.41 & 2.71 & 2.71 & 9.85 & 7.33 \\
GPT 5.4 & 9.40 & 7.22 & 8.09 & 0.43 & 0.41 & 0.44 & 3.77 & 3.85 & 6.91 & 7.44 \\
GPT 5.6 Terra & 6.90 & 7.55 & 6.63 & 0.39 & 0.39 & 0.43 & 2.79 & 2.95 & 6.81 & 5.11 \\
GPT 5.6 Sol & \underline{6.54} & \underline{6.83} & \underline{6.44} & 0.39 & 0.41 & 0.45 & 2.05 & 2.45 & \underline{5.33} & \underline{3.89} \\
Gemini 3.1 Flash & 9.67 & 11.31 & 9.47 & 0.42 & 0.39 & 0.45 & 5.02 & 5.41 & 12.37 & 10.22 \\
Gemini 3.1 Pro & \textbf{5.36} & 8.56 & \textbf{4.21} & 0.41 & 0.41 & 0.44 & 2.13 & 2.43 & \textbf{1.86} & 4.44 \\
\midrule
Socrates-Llama3-8B-DPO & 9.44 & 14.78 & 10.25 & 0.73 & 0.70 & 0.72 & 10.08 & 10.35 & 15.39 & 14.56 \\
Socrates-Qwen-14B-DPO & 10.22 & 15.72 & 10.32 & 0.54 & 0.54 & 0.57 & 7.03 & 7.17 & 12.43 & 13.89 \\
Centaur-70B & 32.31 & 40.36 & 38.05 & 0.78 & 0.77 & 0.87 & 5.24 & 4.86 & 15.73 & 15.89 \\
OSim-8B & 11.89 & 17.05 & 13.78 & 0.53 & 0.53 & 0.62 & 8.06 & 8.33 & 9.70 & 13.89 \\
\bottomrule
\end{tabular}
}
\vspace{5pt}
\caption{Direct distribution elicitation: Wasserstein distance ($\downarrow$) between each model's verbalized distribution and the human distribution, per task. \textbf{Bold} and \underline{underline} denote the best and second best performance. Task names follow Table~\ref{tab:main-results}.}
\label{tab:direct-elicitation}
\end{table*}

%% file: sections/appendix/per-task-results.tex
\section{Per-game Results for Behavior Prediction and Simulation}
\label{app:predict-simulate-details}

This appendix reports the per-game results for the three economic-game tasks summarized in the main results table (Table~\ref{tab:econ-survey-summary}): single-round game behavior simulation, multi-round behavior prediction, and across-game single-round prediction. For each task we report distributional alignment with Wasserstein distance ($W$, lower is better) and individual-level prediction with mean absolute error (MAE, lower is better). The single-round simulation results appear in Table~\ref{tab:game-behavior-wdist}, the multi-round prediction results in Tables~\ref{tab:multiround-wdist} and~\ref{tab:multiround-mae}, and the across-game prediction results in Tables~\ref{tab:acrossgame-wdist} and~\ref{tab:acrossgame-mae}.

\subsection{Economic Games}
\label{app:predict-simulate-econ}

\input{tables/results-game-behavior-wdist}

\input{tables/results-multiround-wdist}

\input{tables/results-multiround-mae}

\input{tables/results-acrossgame-wdist}

\input{tables/results-acrossgame-mae}

\section{Distributions of Model Outputs}
\label{app:dist-plots}

\begin{figure*}[!t]
    \centering
    \includegraphics[width=\linewidth]{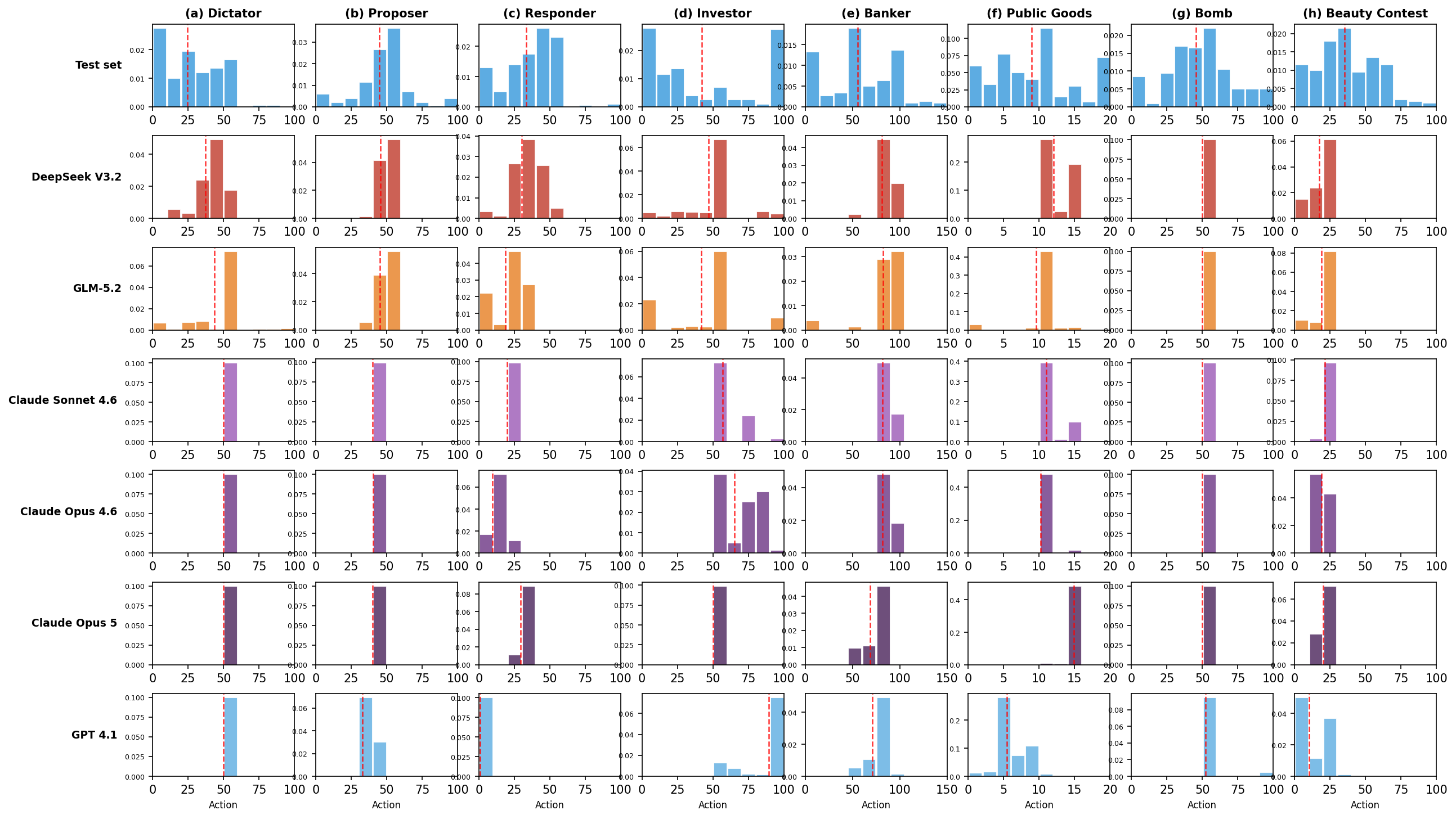}
    \caption{Distribution of model outputs in single-round game behavior simulation (Part 1).}
    \label{fig:game-behavior-dist-part1}
\end{figure*}

\begin{figure*}[!t]
    \centering
    \includegraphics[width=\linewidth]{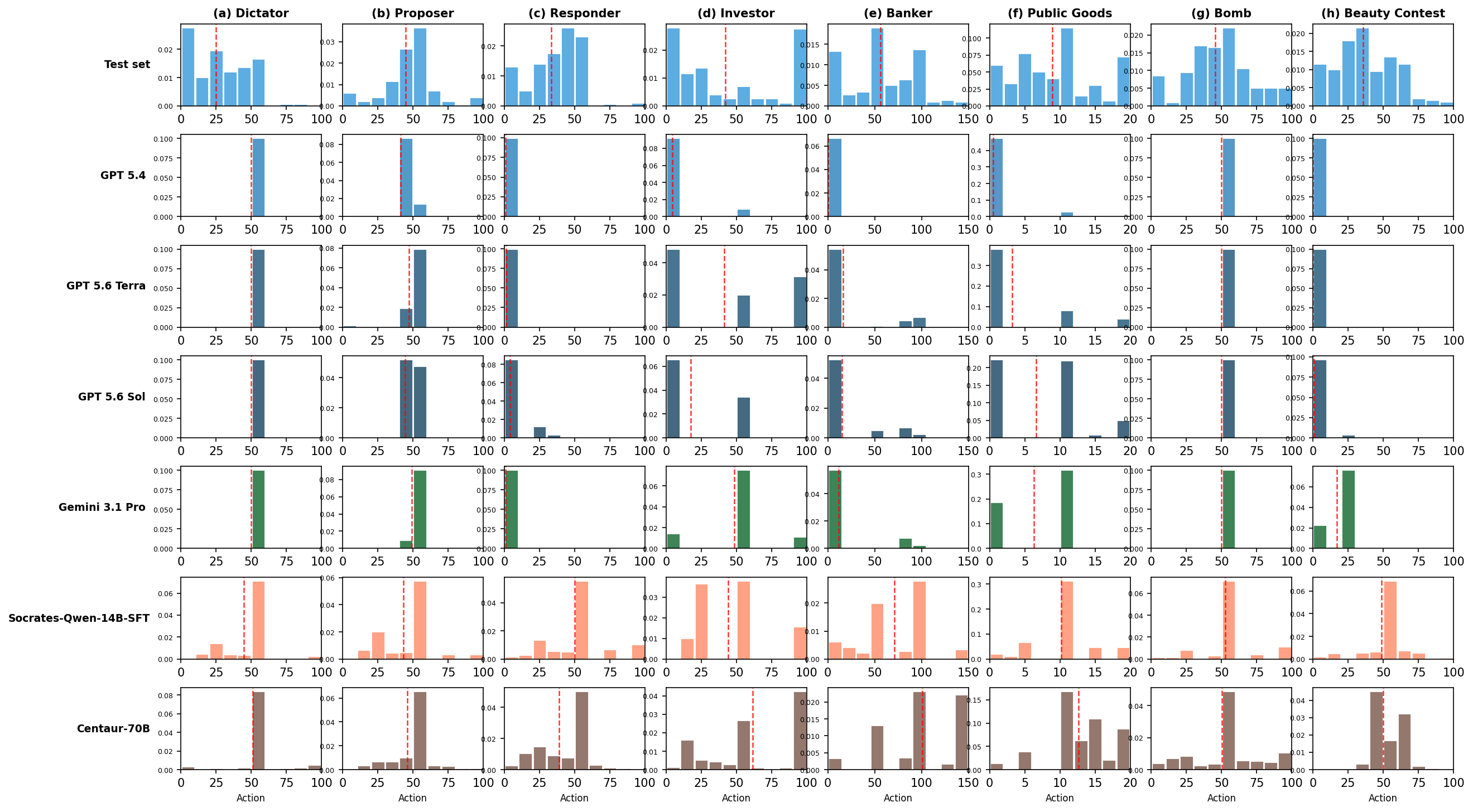}
    \caption{Distribution of model outputs in single-round game behavior simulation (Part 2).}
    \label{fig:game-behavior-dist-part2}
\end{figure*}

\begin{figure*}[!t]
    \centering
    \includegraphics[width=\linewidth]{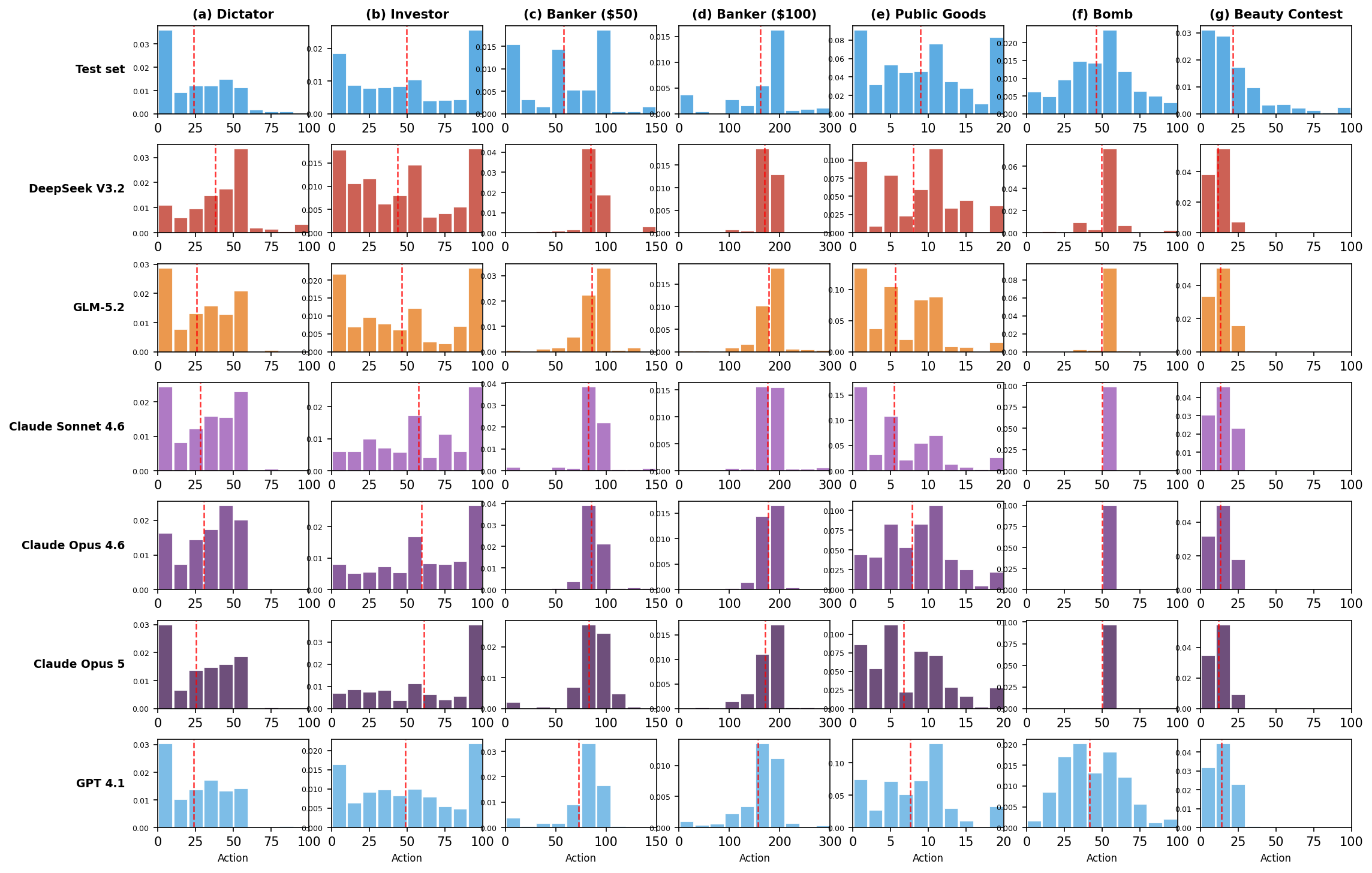}
    \caption{Distribution of model outputs in multi-round game behavior prediction (Part 1).}
    \label{fig:multiround-dist-part1}
\end{figure*}

\begin{figure*}[!t]
    \centering
    \includegraphics[width=\linewidth]{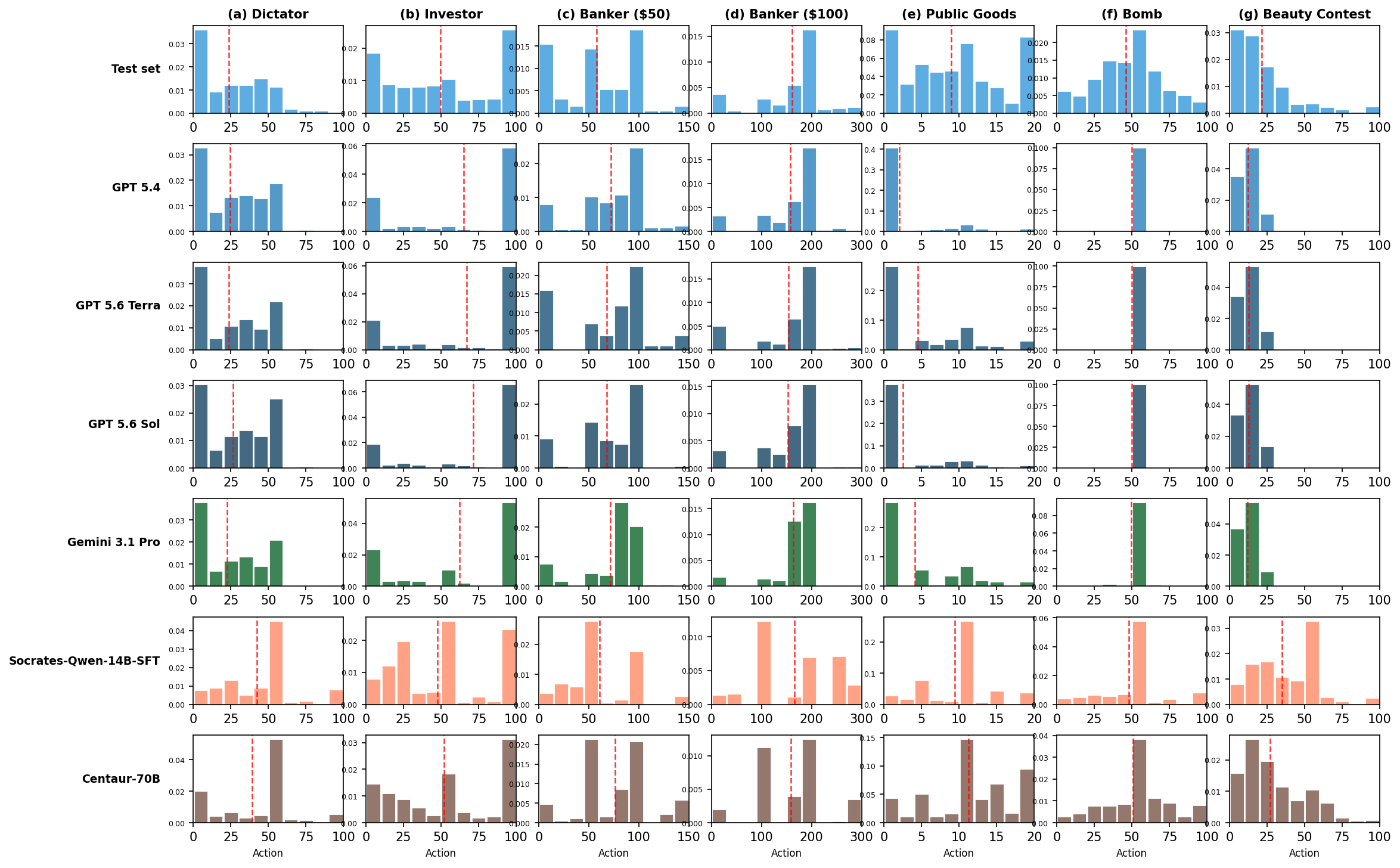}
    \caption{Distribution of model outputs in multi-round game behavior prediction (Part 2).}
    \label{fig:multiround-dist-part2}
\end{figure*}

\begin{figure*}[!t]
    \centering
    \includegraphics[width=\linewidth]{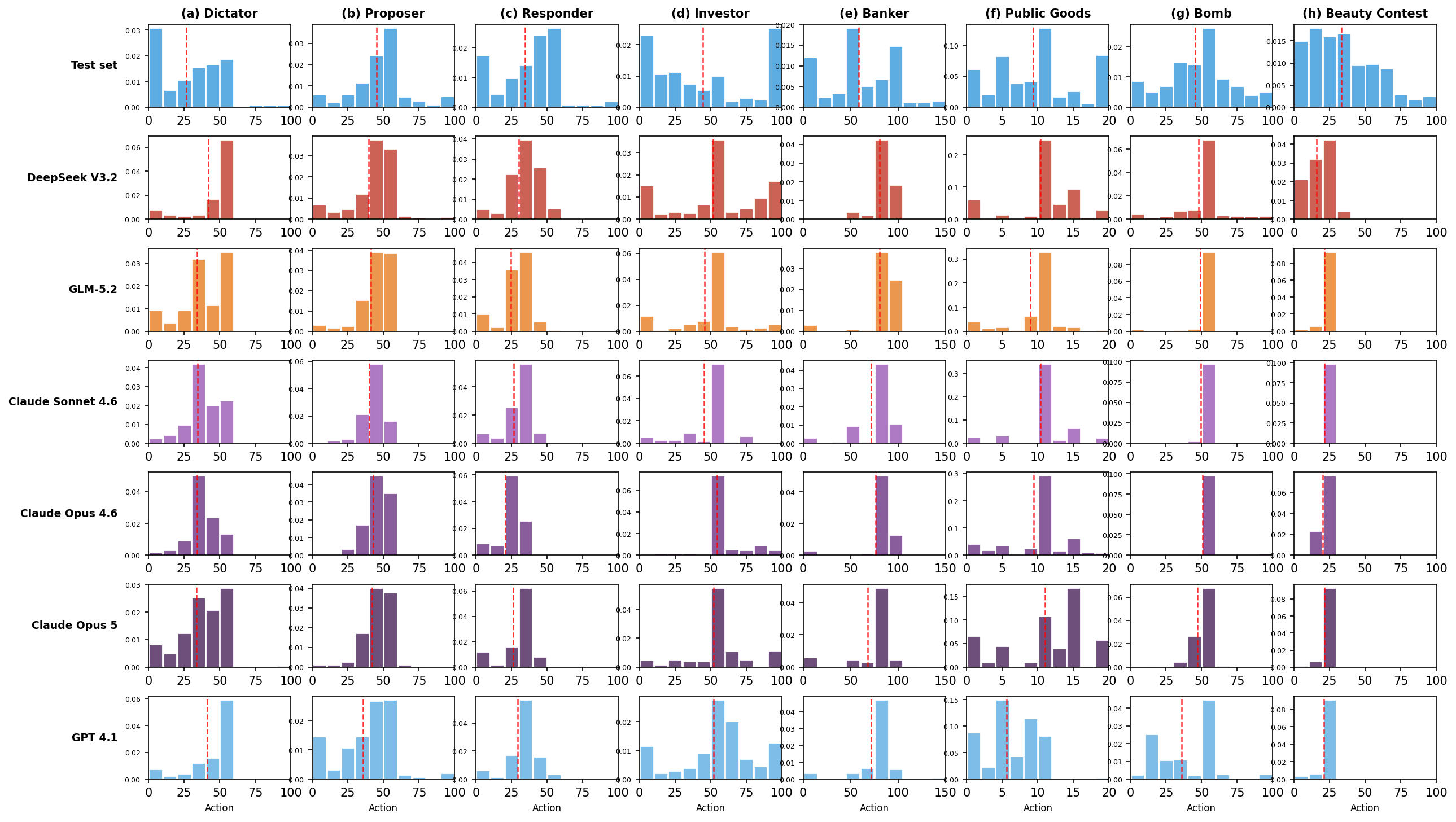}
    \caption{Distribution of model outputs in single-round game behavior prediction given observations from other games (Part 1).}
    \label{fig:acrossgame-dist-part1}
\end{figure*}

\begin{figure*}[!t]
    \centering
    \includegraphics[width=\linewidth]{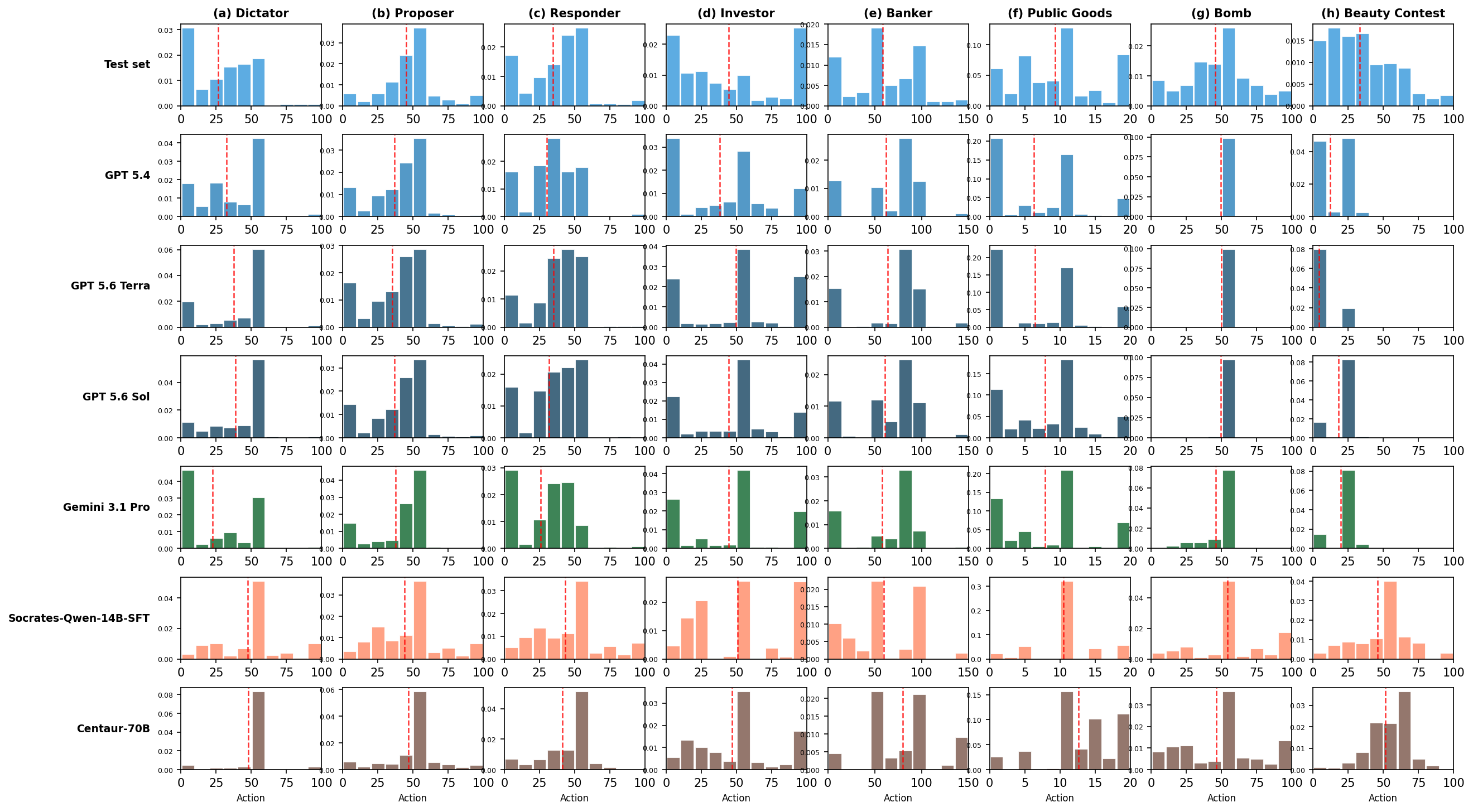}
    \caption{Distribution of model outputs in single-round game behavior prediction given observations from other games (Part 2).}
    \label{fig:acrossgame-dist-part2}
\end{figure*}

This section visualizes the model-predicted output distribution alongside the test-set distribution. The plots cover single-round game behavior simulation (Figures~\ref{fig:game-behavior-dist-part1} and~\ref{fig:game-behavior-dist-part2}), multi-round game behavior prediction (Figures~\ref{fig:multiround-dist-part1} and~\ref{fig:multiround-dist-part2}), and across-game single-round prediction (Figures~\ref{fig:acrossgame-dist-part1} and~\ref{fig:acrossgame-dist-part2}). The top row of each figure shows the test-set distribution, and the remaining rows show the per-model predicted distributions.

\clearpage

%% file: tables/results-game-behavior-wdist.tex
\begin{table*}[!t]
\centering
\resizebox{\textwidth}{!}{%
\small
\begin{tabular}{lccccccccc|c}
\toprule
\textbf{Model} & \textbf{Dictator} & \textbf{Proposer} & \textbf{Responder} & \textbf{Investor} & \textbf{Banker} & \textbf{Pub.Goods} & \textbf{Bomb} & \textbf{Beauty Contest} & \textbf{Push/Pull} & \textbf{Avg} \\
\midrule
Qwen3-4B & 21.3 & 41.0 & 33.6 & 29.0 & 23.2 & 22.8 & 18.9 & 33.9 & 18.2 & 26.9 \\
Llama3.3-70B & 20.1 & 14.8 & 33.4 & 21.3 & 30.2 & 18.8 & 18.6 & 15.3 & 45.3 & 24.2 \\
HY3 & 25.0 & 9.0 & 31.7 & 31.0 & 36.1 & 42.8 & 17.7 & 34.2 & 42.6 & 30.0 \\
DeepSeek V3.2 & \textbf{13.7} & 8.0 & \underline{7.6} & 27.1 & 18.9 & 22.9 & 17.7 & 18.0 & 18.0 & \underline{16.9} \\
GLM-5.2 & \underline{19.0} & 7.6 & 14.8 & 20.3 & 19.3 & 20.2 & 17.7 & 17.0 & 47.7 & 20.4 \\
\midrule
Claude Sonnet 4.6 & 25.8 & 13.6 & 18.7 & 30.8 & 19.1 & 19.5 & 17.7 & 19.5 & 57.0 & 24.6 \\
Claude Opus 4.6 & 25.8 & 13.3 & 23.8 & 33.3 & 19.2 & 23.6 & 17.7 & 20.4 & 57.0 & 26.0 \\
Claude Opus 5 & 25.8 & 13.7 & 14.2 & 37.3 & 18.6 & 37.2 & 17.7 & 19.1 & 57.0 & 26.7 \\
GPT 4.1 & 25.8 & 16.5 & 32.4 & 47.1 & 19.6 & 21.3 & 15.4 & 25.1 & 44.5 & 27.5 \\
GPT 5.4 & 25.8 & 12.3 & 32.4 & 37.9 & 37.1 & 41.8 & 17.7 & 35.5 & 42.0 & 31.4 \\
GPT 5.6 Terra & 25.8 & 9.3 & 32.2 & \underline{7.1} & 26.4 & 28.6 & 17.7 & 35.5 & 42.5 & 25.0 \\
GPT 5.6 Sol & 25.8 & 8.7 & 29.2 & 24.7 & 27.2 & 12.8 & 17.7 & 34.7 & 43.0 & 24.9 \\
Gemini 3.1 Flash & 25.8 & 12.0 & 14.9 & 57.8 & 22.8 & 53.7 & 17.7 & 22.5 & 54.0 & 31.3 \\
Gemini 3.1 Pro & 25.8 & 11.1 & 32.4 & 25.0 & 29.5 & 16.9 & 17.7 & 18.6 & \underline{15.0} & 21.3 \\
\midrule
Socrates-Llama3-8B-SFT & 26.6 & 7.3 & 16.1 & \textbf{5.5} & \textbf{5.6} & \textbf{9.0} & \underline{9.6} & \textbf{10.2} & 57.0$^\dag$ & \textbf{16.3} \\
Socrates-Llama3-8B-DPO & 25.8 & 12.0 & 18.1 & 33.6 & 22.9 & 25.1 & 17.6 & 25.3 & 57.0 & 26.4 \\
Socrates-Qwen-14B-SFT & 19.9 & \textbf{4.4} & 16.7 & 16.1 & \underline{10.1} & \underline{11.6} & 10.4 & 15.3 & 57.0$^\dag$ & 17.9 \\
Socrates-Qwen-14B-DPO & 25.8 & 19.7 & 14.7 & 37.5 & 22.9 & 24.6 & 17.7 & \underline{14.0} & \textbf{3.6} & 20.0 \\
Centaur-70B & 26.2 & \underline{4.8} & \textbf{5.7} & 19.3 & 29.9 & 19.0 & \textbf{5.6} & 15.7 & 26.0 & \underline{16.9} \\
OSim-8B & 22.8 & 7.4 & 32.6 & 37.5 & 16.7 & 18.3 & 13.5 & 17.3 & 30.7 & 21.9 \\
\bottomrule
\end{tabular}
}
\vspace{5pt}
\caption{Single-round game behavior simulation: Wasserstein distance ($\downarrow$). \textbf{Bold} and \underline{underline} denote the best and second best performance. $^\dag$ The model produced no parseable Push/Pull response, so its Push/Pull $W$ is set to the worst observed value in this column.}
\label{tab:game-behavior-wdist}
\end{table*}

%% file: tables/results-multiround-wdist.tex
\begin{table*}[!t]
\centering
\resizebox{\textwidth}{!}{%
\small
\begin{tabular}{lcccccccc|c}
\toprule
\textbf{Model} & \textbf{Dictator} & \textbf{Investor} & \textbf{Banker\_inv50} & \textbf{Banker\_inv100} & \textbf{Pub.Goods} & \textbf{Bomb} & \textbf{Beauty Contest} & \textbf{Push/Pull} & \textbf{Avg} \\
\midrule
Qwen3-4B & 13.5 & 22.2 & 19.3 & 7.7 & 15.0 & 13.6 & 9.2 & 4.0 & 13.1 \\
Llama3.3-70B & 21.8 & 12.6 & 27.3 & 13.8 & 15.0 & 12.0 & 7.3 & 12.8 & 15.3 \\
HY3 & 2.2 & 14.8 & \textbf{5.1} & 3.7 & 42.5 & 16.0 & 9.0 & 33.0 & 15.8 \\
DeepSeek V3.2 & 14.2 & 6.1 & 18.0 & 11.4 & \textbf{5.5} & 12.7 & 9.9 & \underline{3.0} & \underline{10.1} \\
GLM-5.2 & 3.6 & 3.2 & 19.9 & 7.9 & 16.8 & 16.2 & 8.9 & 4.4 & \underline{10.1} \\
\midrule
Claude Sonnet 4.6 & 5.9 & 8.3 & 16.9 & 9.4 & 17.5 & 17.2 & 8.7 & \textbf{0.8} & 10.6 \\
Claude Opus 4.6 & 8.5 & 9.8 & 19.3 & 9.5 & 9.8 & 17.3 & 8.6 & 10.4 & 11.7 \\
Claude Opus 5 & 3.2 & 11.6 & 18.1 & 9.0 & 11.5 & 16.8 & 9.6 & 12.6 & 11.5 \\
GPT 4.1 & 2.5 & \textbf{2.3} & 15.1 & 8.4 & 8.4 & \underline{5.0} & 8.1 & 5.6 & \textbf{6.9} \\
GPT 5.4 & 2.1 & 17.3 & 9.7 & \textbf{2.5} & 34.7 & 17.2 & 9.4 & 28.6 & 15.2 \\
GPT 5.6 Terra & \underline{2.0} & 18.4 & 7.3 & \underline{2.6} & 22.3 & 17.3 & 9.2 & 32.8 & 14.0 \\
GPT 5.6 Sol & 4.2 & 22.5 & 8.2 & 3.8 & 32.5 & 17.3 & 9.0 & 31.2 & 16.1 \\
Gemini 3.1 Flash & 8.6 & 27.2 & 19.1 & 9.1 & \underline{7.4} & 16.1 & 10.3 & 50.8 & 18.6 \\
Gemini 3.1 Pro & \textbf{1.6} & 14.7 & 10.9 & 7.1 & 24.4 & 16.4 & 9.7 & 15.0 & 12.5 \\
\midrule
Socrates-Llama3-8B-SFT & 15.6 & 15.8 & 9.1 & 29.3 & 19.6 & 11.3 & 19.0 & 50.8$^\dag$ & 21.3 \\
Socrates-Llama3-8B-DPO & 26.5 & 12.0 & 22.5 & 14.0 & 21.4 & 9.4 & 11.6 & 40.1 & 19.7 \\
Socrates-Qwen-14B-SFT & 18.5 & 5.5 & \underline{6.9} & 11.5 & 12.8 & 6.8 & 14.0 & 50.8$^\dag$ & 15.9 \\
Socrates-Qwen-14B-DPO & 14.5 & 15.2 & 20.5 & 14.1 & 10.8 & 7.8 & \textbf{6.6} & 26.0 & 14.4 \\
Centaur-70B & 15.3 & \underline{2.6} & 12.3 & 8.1 & 11.5 & \underline{5.0} & \underline{6.8} & 27.0 & 11.1 \\
OSim-8B & 16.5 & 14.9 & 20.5 & 9.9 & 14.1 & \textbf{3.6} & 7.9 & 12.0 & 12.4 \\
\bottomrule
\end{tabular}
}
\vspace{5pt}
\caption{Multi-round game behavior prediction: Wasserstein distance ($\downarrow$). \textbf{Bold} and \underline{underline} denote the best and second best performance. $^\dag$ The model produced no parseable Push/Pull response, so its Push/Pull $W$ is set to the worst observed value in this column.}
\label{tab:multiround-wdist}
\end{table*}

%% file: tables/results-multiround-mae.tex
\begin{table*}[!t]
\centering
\resizebox{\textwidth}{!}{%
\small
\begin{tabular}{lcccccccc|c}
\toprule
\textbf{Model} & \textbf{Dictator} & \textbf{Investor} & \textbf{Banker\_inv50} & \textbf{Banker\_inv100} & \textbf{Pub.Goods} & \textbf{Bomb} & \textbf{Beauty Contest} & \textbf{Push/Pull} & \textbf{Avg} \\
\midrule
Qwen3-4B & 18.7 & 30.0 & 23.5 & 17.0 & 20.9 & 17.8 & 11.4 & 34.0 & 21.7 \\
Llama3.3-70B & 23.8 & 35.4 & 28.8 & 18.4 & 26.6 & 18.5 & 11.1 & 53.6 & 27.0 \\
HY3 & 10.1 & 33.1 & 17.5 & 15.6 & 42.9 & 17.6 & 11.5 & 35.4 & 23.0 \\
DeepSeek V3.2 & 20.2 & 25.0 & 24.9 & 16.0 & 22.5 & 17.4 & 11.6 & 29.4 & 20.9 \\
GLM-5.2 & 10.3 & 23.9 & 22.4 & \textbf{12.0} & 24.3 & \underline{16.5} & 11.1 & 30.4 & 18.8 \\
\midrule
Claude Sonnet 4.6 & 11.8 & \underline{22.7} & 20.7 & 13.2 & 24.5 & 17.2 & 11.0 & 28.8 & 18.7 \\
Claude Opus 4.6 & 13.0 & \underline{22.7} & 22.8 & 13.2 & 21.7 & 17.3 & 11.0 & 32.4 & 19.3 \\
Claude Opus 5 & \underline{9.7} & 26.4 & 21.5 & \underline{12.8} & 20.8 & 17.0 & 11.2 & \textbf{22.6} & \textbf{17.7} \\
GPT 4.1 & 10.1 & \textbf{21.0} & 21.3 & 15.0 & \textbf{18.3} & \textbf{16.3} & \underline{10.6} & 32.4 & \underline{18.1} \\
GPT 5.4 & \textbf{9.4} & 33.0 & \underline{16.9} & 13.7 & 36.4 & 17.2 & 11.2 & 31.4 & 21.1 \\
GPT 5.6 Terra & 10.3 & 33.9 & 18.9 & 16.2 & 29.1 & 17.3 & 11.2 & 36.8 & 21.7 \\
GPT 5.6 Sol & 11.0 & 35.8 & \textbf{16.2} & 14.2 & 35.6 & 17.3 & 11.0 & 34.8 & 22.0 \\
Gemini 3.1 Flash & 14.2 & 36.3 & 24.2 & 14.1 & 26.2 & 17.9 & 11.7 & 59.2 & 25.5 \\
Gemini 3.1 Pro & \underline{9.7} & 31.3 & 19.2 & 13.7 & 29.7 & 16.8 & 11.5 & \underline{23.4} & 19.4 \\
\midrule
Socrates-Llama3-8B-SFT & 25.9 & 37.6 & 28.1 & 36.9 & 32.1 & 28.1 & 27.7 & 100.0$^\dag$ & 39.5 \\
Socrates-Llama3-8B-DPO & 27.9 & 26.5 & 23.9 & 17.8 & 26.9 & 18.4 & 18.7 & 57.8 & 27.3 \\
Socrates-Qwen-14B-SFT & 27.9 & 33.0 & 26.1 & 27.7 & 26.5 & 22.5 & 23.7 & 100.0$^\dag$ & 35.9 \\
Socrates-Qwen-14B-DPO & 18.4 & 26.4 & 25.1 & 17.0 & \underline{19.5} & \underline{16.5} & \textbf{10.0} & 30.8 & 20.5 \\
Centaur-70B & 19.8 & 23.5 & 26.6 & 22.6 & 24.7 & 19.0 & 15.5 & 34.6 & 23.3 \\
OSim-8B & 21.0 & 27.1 & 27.1 & 20.3 & 22.0 & 17.5 & 10.9 & 35.6 & 22.7 \\
\bottomrule
\end{tabular}
}
\vspace{5pt}
\caption{Multi-round game behavior prediction: MAE ($\downarrow$). \textbf{Bold} and \underline{underline} denote the best and second best performance. $^\dag$ The model produced no parseable Push/Pull response; all such cases are counted as incorrect predictions.}
\label{tab:multiround-mae}
\end{table*}

%% file: tables/results-acrossgame-wdist.tex
\begin{table*}[!t]
\centering
\resizebox{\textwidth}{!}{%
\small
\begin{tabular}{lccccccccc|c}
\toprule
\textbf{Model} & \textbf{Dictator} & \textbf{Proposer} & \textbf{Responder} & \textbf{Investor} & \textbf{Banker} & \textbf{Pub.Goods} & \textbf{Bomb} & \textbf{Beauty Contest} & \textbf{Push/Pull} & \textbf{Avg} \\
\midrule
Qwen3-4B & 16.3 & 21.7 & 28.4 & 27.1 & 19.7 & 22.0 & 17.7 & 18.3 & \textbf{9.5} & 20.1 \\
Llama3.3-70B & 14.5 & 9.5 & 10.4 & 20.7 & 12.4 & 16.7 & 16.8 & 14.0 & 18.7 & 14.9 \\
HY3 & 9.2 & 9.1 & 6.8 & 10.0 & 6.4 & 19.9 & 16.6 & 15.7 & 32.3 & 14.0 \\
DeepSeek V3.2 & 16.2 & 5.4 & 9.8 & 11.4 & 16.6 & 11.1 & 10.0 & 17.3 & \underline{12.9} & \underline{12.3} \\
GLM-5.2 & 8.7 & 6.6 & 13.2 & 22.0 & 17.0 & 15.6 & 16.6 & 18.4 & 50.1 & 18.7 \\
\midrule
Claude Sonnet 4.6 & 9.9 & 9.6 & 12.1 & 26.0 & 14.8 & 14.4 & 17.3 & 19.0 & 50.7 & 19.3 \\
Claude Opus 4.6 & 11.4 & 7.9 & 16.5 & 26.8 & 16.6 & 12.0 & 16.7 & 18.7 & 54.3 & 20.1 \\
Claude Opus 5 & 8.1 & 7.5 & 11.6 & 21.0 & 15.0 & 12.0 & 14.6 & 18.9 & 29.7 & 15.4 \\
GPT 4.1 & 15.4 & 9.4 & 11.0 & 16.6 & 16.0 & 18.4 & 10.3 & 18.3 & 30.9 & 16.3 \\
GPT 5.4 & \underline{5.9} & 8.4 & 5.2 & 10.4 & 6.0 & 14.9 & 17.4 & 21.1 & 37.2 & 14.0 \\
GPT 5.6 Terra & 11.2 & 10.0 & \underline{4.0} & 8.2 & 6.8 & 14.1 & 17.6 & 29.0 & 33.1 & 14.9 \\
GPT 5.6 Sol & 12.4 & 8.3 & \textbf{3.4} & 12.5 & \underline{5.7} & \textbf{7.5} & 17.2 & 16.9 & 36.3 & 13.4 \\
Gemini 3.1 Flash & 19.9 & 7.1 & 14.3 & 11.9 & 14.7 & 12.4 & 13.6 & 20.5 & 41.1 & 17.3 \\
Gemini 3.1 Pro & \textbf{5.2} & 7.4 & 8.9 & 9.7 & 8.0 & \underline{7.9} & 13.5 & 16.7 & 15.3 & \textbf{10.3} \\
\midrule
Socrates-Llama3-8B-SFT & 21.8 & 6.4 & 8.3 & \underline{7.9} & 7.6 & 8.1 & \underline{8.3} & \textbf{8.7} & 54.3$^\dag$ & 14.6 \\
Socrates-Llama3-8B-DPO & 23.7 & 7.5 & 5.6 & 23.7 & 21.1 & 26.1 & 9.8 & 28.1 & 46.9 & 21.4 \\
Socrates-Qwen-14B-SFT & 20.8 & 4.8 & 8.8 & \textbf{6.7} & \textbf{4.0} & 11.1 & 9.0 & \underline{13.1} & 54.3$^\dag$ & 14.7 \\
Socrates-Qwen-14B-DPO & 20.1 & \underline{3.5} & 11.8 & 31.6 & 19.5 & 21.4 & 17.7 & 15.3 & 18.8 & 17.7 \\
Centaur-70B & 21.4 & \textbf{2.9} & 7.6 & 11.4 & 14.4 & 17.2 & \textbf{5.3} & 19.8 & 24.8 & 13.9 \\
OSim-8B & 11.0 & 13.0 & 33.6 & 29.0 & 16.0 & 17.6 & 17.6 & 16.2 & 37.1 & 21.2 \\
\bottomrule
\end{tabular}
}
\vspace{5pt}
\caption{Single-round game behavior prediction given observations from other games: Wasserstein distance ($\downarrow$). \textbf{Bold} and \underline{underline} denote the best and second best performance. $^\dag$ The model produced no parseable Push/Pull response, so its Push/Pull $W$ is set to the worst observed value in this column.}
\label{tab:acrossgame-wdist}
\end{table*}

%% file: tables/results-acrossgame-mae.tex
\begin{table*}[!t]
\centering
\resizebox{\textwidth}{!}{%
\small
\begin{tabular}{lccccccccc|c}
\toprule
\textbf{Model} & \textbf{Dictator} & \textbf{Proposer} & \textbf{Responder} & \textbf{Investor} & \textbf{Banker} & \textbf{Pub.Goods} & \textbf{Bomb} & \textbf{Beauty Contest} & \textbf{Push/Pull} & \textbf{Avg} \\
\midrule
Qwen3-4B & 22.0 & 24.4 & 29.9 & \textbf{32.7} & 26.2 & \underline{24.7} & \textbf{17.7} & 26.1 & 47.3 & 27.9 \\
Llama3.3-70B & 24.4 & 15.0 & 18.3 & 36.7 & 26.9 & 25.7 & 18.2 & 23.4 & 48.8 & 26.4 \\
HY3 & 24.3 & 14.7 & 16.8 & 39.5 & 26.5 & 35.3 & 18.1 & 22.6 & 46.4 & 27.1 \\
DeepSeek V3.2 & 24.3 & 13.6 & 16.5 & 40.2 & 24.5 & 30.2 & 21.9 & 23.1 & 52.9 & 27.5 \\
GLM-5.2 & 21.6 & 13.3 & 16.8 & 36.0 & 25.6 & 25.9 & 18.1 & 19.9 & 54.4 & 25.7 \\
\midrule
Claude Sonnet 4.6 & \underline{20.7} & \underline{13.1} & 15.7 & 35.9 & \underline{22.8} & 26.5 & \underline{17.9} & \underline{19.8} & 55.2 & \underline{25.3} \\
Claude Opus 4.6 & \textbf{19.4} & \textbf{12.3} & 19.0 & 35.4 & 24.4 & 27.7 & 18.2 & 20.2 & 56.1 & 25.9 \\
Claude Opus 5 & 21.2 & \textbf{12.3} & 15.6 & 36.5 & 23.4 & 33.4 & 18.2 & 19.9 & 51.9 & 25.8 \\
GPT 4.1 & 22.6 & 15.4 & 15.4 & 39.3 & 23.9 & 29.2 & 26.2 & 20.2 & 51.5 & 27.1 \\
GPT 5.4 & 22.5 & 14.1 & 14.7 & 37.2 & 25.5 & 32.5 & \underline{17.9} & 25.9 & \underline{44.7} & 26.1 \\
GPT 5.6 Terra & 23.8 & 15.3 & \textbf{12.4} & 40.9 & 24.6 & 34.7 & \textbf{17.7} & 30.4 & 45.1 & 27.2 \\
GPT 5.6 Sol & 23.8 & 14.1 & \underline{14.1} & 38.7 & 24.5 & 30.6 & \underline{17.9} & 22.0 & \textbf{44.0} & 25.5 \\
Gemini 3.1 Flash & 24.8 & 14.0 & 20.0 & 45.5 & 23.9 & 37.2 & 21.7 & 24.0 & 53.6 & 29.4 \\
Gemini 3.1 Pro & 23.8 & 14.1 & 17.9 & 39.1 & 25.3 & 32.2 & 19.3 & 21.6 & 47.1 & 26.7 \\
\midrule
Socrates-Llama3-8B-SFT & 28.7 & 23.4 & 25.5 & 39.6 & 24.8 & 39.6 & 30.8 & 26.8 & 100.0$^\dag$ & 37.7 \\
Socrates-Llama3-8B-DPO & 25.0 & 14.6 & 16.5 & \underline{34.0} & \textbf{21.7} & 28.2 & 21.7 & 30.7 & 56.8 & 27.7 \\
Socrates-Qwen-14B-SFT & 29.7 & 22.8 & 24.3 & 42.2 & 26.4 & 30.1 & 27.8 & 26.6 & 100.0$^\dag$ & 36.7 \\
Socrates-Qwen-14B-DPO & 23.7 & 16.0 & 16.5 & 34.4 & 22.9 & \textbf{24.3} & \textbf{17.7} & \textbf{19.7} & 48.1 & \textbf{24.8} \\
Centaur-70B & 26.0 & 16.3 & 15.8 & 37.5 & 30.3 & 34.8 & 29.5 & 27.9 & 45.6 & 29.3 \\
OSim-8B & 24.0 & 21.3 & 33.6 & 35.6 & 24.7 & 26.3 & 33.6 & 20.4 & 45.3 & 29.4 \\
\bottomrule
\end{tabular}
}
\vspace{5pt}
\caption{Single-round game behavior prediction given observations from other games: MAE ($\downarrow$). \textbf{Bold} and \underline{underline} denote the best and second best performance. $^\dag$ The model produced no parseable Push/Pull response; all such cases are counted as incorrect predictions.}
\label{tab:acrossgame-mae}
\end{table*}

%% file: sections/appendix/prompt-templates.tex
\section{Prompt Templates}
\label{app:prompt-templates}

\input{tables/prompt-templates}

%% file: tables/prompt-templates.tex
\mdfdefinestyle{promptstyle}{%
  linewidth=0.5pt,
  innerleftmargin=8pt,
  innerrightmargin=8pt,
  innertopmargin=6pt,
  innerbottommargin=6pt,
  skipabove=4pt,
  skipbelow=4pt,
}

\newcommand{\ph}[1]{\{{\normalfont\itshape #1}\}}

This appendix documents all prompt templates used in the experiments.
For each task, we show the system prompt and user prompt template.
Placeholders are shown as \ph{variable}.
Section headers within prompts (originally in Markdown) are rendered in \textbf{bold}.

\begin{small}

\phantomsection\label{prompt:games-firstround}
\subsection*{A.\quad Economic Games --- Single-round Behavior Simulation}

All single-round economic game prompts share the following system prompt.

\paragraph{System prompt.}
\begin{mdframed}[style=promptstyle]
You are a player who is playing an economics game.
\end{mdframed}

\phantomsection\label{prompt:dictator}
\paragraph{Dictator Game.}
\begin{mdframed}[style=promptstyle]
You are paired with another player. Your role is to decide how to divide \$100
and the other player simply receives your choice. How would you like to divide
the money? Please give only one concrete choice and highlight the amount you
give to the other player in [] (such as [\$x]).
\end{mdframed}

\phantomsection\label{prompt:ultimatum-proposer}
\paragraph{Ultimatum Game --- Proposer.}
\begin{mdframed}[style=promptstyle]
This is a two-player game. You are the Proposer, and the other player is the
Responder. As the proposer, you propose how to divide \$100 and the Responder
chooses either Accept or Reject. If accepted, the two of you will earn as
described by the accepted proposal accordingly. If rejected, then both of you
will earn \$0.
How much would you like to propose to give to the Responder? Please give only
one concrete choice, and highlight the amount with [] (such as [\$x]).
\end{mdframed}

\phantomsection\label{prompt:ultimatum-responder}
\paragraph{Ultimatum Game --- Responder.}
\begin{mdframed}[style=promptstyle]
This is a two-player game. You are the Responder, and the other player is the
Proposer. The proposer proposes how to divide \$100 and you, as the Responder,
choose either Accept or Reject. If accepted, the two of you will earn as
described by the accepted proposal accordingly. If rejected, then both of you
will earn \$0.
As the Responder, what is the minimal amount in the proposal that you would
accept? Please give only one concrete choice, and highlight the amount with []
(such as [\$x]).
\end{mdframed}

\phantomsection\label{prompt:trust-investor}
\paragraph{Trust Game --- Investor.}
\begin{mdframed}[style=promptstyle]
This is a two-player game. You are an Investor and the other player is a Banker.
You have \$100 to invest and you choose how much of your money to invest with the
Banker. The amount you choose to invest will grow by 3x with the Banker. For
example, if you invest \$10, it will grow to \$30 with the Banker. The Banker then
decides how much of the money (\$0-\$30) to return to you, the Investor.
How much of the \$100 would you like to invest with the Banker? Please give only
one concrete choice, and highlight the number with [] (such as [\$x]).
\end{mdframed}

\phantomsection\label{prompt:trust-banker}
\paragraph{Trust Game --- Banker.}
\begin{mdframed}[style=promptstyle]
This is a two-player game. You are a Banker and the other player is an Investor,
and the goal for each player is to earn more. The Investor chooses how much of
the money (up to \$100) to invest with you. The amount the Investor invests will
generate a 2x return with you (the current value of investment becomes 3x). To
settle the investment, as the Banker, you get to decide how much of this total
amount to return to the Investor and the rest will be kept as your profit. For
example, you can choose to return \$0 (therefore the investor will lose their
investment), or you can return the entire 3x (initial investment + 2x profit) to
the investor, or you can return any amount in between.
Now, if the investor has invested \$50 with you and the current value became \$150,
how much of the \$150 would you like to return to the Investor? Please give only
one concrete choice, and highlight the number with [] (such as [\$x]).
\end{mdframed}

\phantomsection\label{prompt:public-goods}
\paragraph{Public Goods Game.}
\begin{mdframed}[style=promptstyle]
In this public good game, you and 3 others will choose how much to contribute
towards a water cleaning project. Each of you gets \$20 per round to contribute
between \$0 and \$20. The project has a 50\% return rate. Your payoff relies on
both your and others' contributions. Everyone benefits from the group's total
contribution. Your payoff in each round equals the amount you didn't contribute
(endowment - contribution) plus everyone's benefit (total contributions * 50\%
return rate). Here are two examples to calculate your payoff.

Example one: You contributed \$12; total group contributions were \$20

Your Payoff = (\$20-\$12) + \$20*50\% = \$18

Example two: You contributed \$12; total group contributions were \$30

Your Payoff = (\$20-\$12) + \$30*50\% = \$23

We will play a total of 3 rounds, in the first round, how much of the \$20 would
you like to contribute? Please give a concrete number and highlight it with []
(e.g., [x]).
\end{mdframed}

\phantomsection\label{prompt:bomb-risk}
\paragraph{Bomb Risk Game.}
\begin{mdframed}[style=promptstyle]
There are 100 boxes, and one bomb has been randomly placed in 1 of 100 boxes.
You can choose to open 0-100 boxes at the same time. If none of the boxes you
open has the bomb, you earn points that are equal to the number of boxes you
open. If one of the boxes you open has the bomb, you earn zero points. How many
boxes would you open? Please give one concrete number and highlight it with []
(such as [x]).
\end{mdframed}

\phantomsection\label{prompt:beauty-contest}
\paragraph{Beauty Contest (Guessing) Game.}
\begin{mdframed}[style=promptstyle]
This is a number-guessing game where you will play with \ph{num\_players} other
players. Each player chooses an integer between 0 and 100 which is hidden from
other players. The player who chooses a number that is closest to 2/3 of the
average wins the game. For example, if the average of the guesses is 60 and if
you guess 40, i.e.\ 2/3 of 60, you win the game. What number would you like to
guess? Please give a concrete number and highlight it with [] (e.g., [x]).
\end{mdframed}

\phantomsection\label{prompt:pushpull-firstround}
\paragraph{Push/Pull (Prisoner's Dilemma) Game.}
\begin{mdframed}[style=promptstyle]
You're paired with another player, each having a \$400 `Push' card and a \$300 `Pull' card. Your payoff depends on both players' card choices. Here are the scenarios:
\begin{itemize}[nosep]
\item Both play `Push': Each earns \$400
\item You play `Push', the other player plays `Pull': You earn \$0, the other player earns \$700
\item You play `Pull', the other player plays `Push': You earn \$700, the other player earns \$0
\item Both play `Pull': Each earns \$300
\end{itemize}
In 5 total rounds, which card would you play in the first round? Please give a concrete choice and highlight it with [] (i.e., [Push] or [Pull]).

\medskip
\noindent\textit{Instruction suffix:}\\
Important: Answer with [Push] or [Pull], not a number.
\end{mdframed}

\phantomsection\label{prompt:games-multiround}
\subsection*{B.\quad Economic Games --- Multi-Round Behavior Prediction}

The system prompt is the same as in Section~A.
Multi-round prompts predict a player's next action given their play history.
The general user prompt template is:

\paragraph{General template.}
\begin{mdframed}[style=promptstyle]
\ph{game\_instruction}

\textbf{Play History}\\
In this session, you have already played \ph{num\_rounds} round(s). Here is the play history:\\
\ph{history}

\textbf{Your Choice}\\
Now in Round \ph{next\_round}, what would you choose? Please give only one concrete choice and
highlight the number with [] (such as [\$x]).
\end{mdframed}

\noindent
Here \ph{game\_instruction} is the single-round prompt from Section~A and
\ph{history} consists of per-round lines in one of the following formats.

\paragraph{History line formats.}
\begin{mdframed}[style=promptstyle]
\textit{Without other players (Dictator, Bomb):}\\
- Round \ph{round\_num}. Your choice: [\ph{your\_action}].

\medskip
\textit{With one other player --- singular (Trust):}\\
- Round \ph{round\_num}. Your choice: [\ph{your\_action}], the other player's choice: [\ph{others\_action}].

\medskip
\textit{With multiple other players --- plural (Public Goods):}\\
- Round \ph{round\_num}. Your choice: [\ph{your\_action}], the other players' choices: [\ph{others\_action}].

\medskip
\textit{Beauty Contest (custom):}\\
- Round \ph{round\_num}. Your choice: [\ph{your\_action}], the group average is [\ph{group\_avg}], and 2/3 of that is [\ph{target}]. \ph{round\_outcome}
\end{mdframed}

\phantomsection\label{prompt:trust-banker-multiround}
\paragraph{Trust Banker --- variable investment.}
In multi-round Trust Banker prompts, the investment amount varies per round.
The game rules (same as Section~A) are followed by the play history and a
round-specific question:
\begin{mdframed}[style=promptstyle]
Now in this round, the investor has invested \$\ph{investment} with you and the
current value became \$\ph{total}. How much of the \$\ph{total} would you like
to return to the Investor? Please give only one concrete choice and highlight
the number with [] (such as [\$x]).
\end{mdframed}

\phantomsection\label{prompt:beauty-contest-multiround}
\paragraph{Beauty Contest --- multi-round.}
The guessing game uses simplified rules (without the round count) and a modified question:
\begin{mdframed}[style=promptstyle]
Now you will play the game again (Round \ph{next\_round}). What number would you
like to guess? Please give a concrete number and highlight it with [] (e.g., [x]).
\end{mdframed}

\phantomsection\label{prompt:pushpull-multiround}
\paragraph{Push/Pull --- multi-round.}
The Push/Pull game uses [Push]/[Pull] choices instead of numeric values:
\begin{mdframed}[style=promptstyle]
Which card would you play in Round \ph{next\_round}? Please give a concrete choice and highlight it with [] (i.e., [Push] or [Pull]).

Respond with only [Push] or [Pull]. Do not explain.
\end{mdframed}

\phantomsection\label{prompt:games-acrossctx}
\subsection*{C.\quad Economic Games --- Across-context Single-round Prediction}

The system prompt is the same as in Section~A.
Given a player's history from other games, the model predicts their choice in a new target game.

\paragraph{User prompt template.}
\begin{mdframed}[style=promptstyle]
\textbf{Play History}\\
You have played the following economics games before.

\ph{history\_section}

\textbf{Your Choice}\\
Now you are playing a new game with the game instruction below:

\ph{target\_instruction}
\end{mdframed}

\noindent
Each entry in \ph{history\_section} is formatted as:

\begin{mdframed}[style=promptstyle]
\ph{index}. \ph{game\_name}\\
- Game instruction: \ph{instruction}\\
- Your choice: [\ph{choice}]
\end{mdframed}

\noindent
The \ph{target\_instruction} is the single-round prompt from Section~A for the target game.

\phantomsection\label{prompt:bigfive}
\subsection*{D.\quad Big Five Personality Survey}

\phantomsection\label{prompt:bigfive-dim}
\paragraph{Task 1: Personality Score from Demographics.}
\begin{mdframed}[style=promptstyle]
\textbf{System:} You are an expert in psychology. Given a person's demographics, your task is to predict this person's BigFive dimensionality scores.

\medskip
\textbf{User:}\\
\textbf{Demographics}\\
\ph{demographics}

\textbf{BigFive Dimensionality Scores}\\
Each dimensionality score ranges from 10 to 50, with 10 indicating the lowest score in that dimension and 50 indicating the highest score.

\textbf{Output Format}\\
Based on this person's demographics, please estimate this person's personality score in the *\ph{dimension\_name}* dimension. Please output a single number in the range from 10 to 50, highlighted in [] (e.g., [x]).
\end{mdframed}

\noindent
\ph{demographics} is formatted as: ``A \ph{age}-year-old \ph{gender} from \ph{country}. \ph{hand}-handed. The race is \ph{race}. The native language is \ph{engnat}.''
\ph{dimension\_name} is one of: Extraversion, Neuroticism, Agreeableness, Conscientiousness, Openness.

\phantomsection\label{prompt:bigfive-singleresp}
\paragraph{Task 2: Survey Response Prediction Given Demographics.}
\begin{mdframed}[style=promptstyle]
\textbf{System:} You are a \ph{age}-year-old \ph{gender} from \ph{country}. You are \ph{hand}-handed. Your race is \ph{race}. Your native language is \ph{engnat}.

\medskip
\textbf{User:} The following item was rated on a five-point scale where 1=Disagree, 2=Partially Disagree, 3=Neutral, 4=Partially Agree, 5=Agree. Please select how this statement describes you and highlight your answer in [](such as [1],[2],[3],[4],or [5]): \ph{question} Only output your answer in brackets.
\end{mdframed}

\phantomsection\label{prompt:bigfive-age}
\paragraph{Task 3: Age Prediction from Personality.}
\begin{mdframed}[style=promptstyle]
\textbf{System:} You are an expert in psychology. Given a person's BigFive personality test dimensionality scores, your task is to predict this person's age.

\medskip
\textbf{User:}\\
\textbf{BigFive Dimensionality Scores}\\
Each dimensionality score ranges from 10 to 50, with 10 indicating the lowest score in that dimension and 50 indicating the highest score.
\begin{itemize}[nosep]
\item Openness: \ph{O\_score}
\item Conscientiousness: \ph{C\_score}
\item Extraversion: \ph{E\_score}
\item Agreeableness: \ph{A\_score}
\item Neuroticism: \ph{N\_score}
\end{itemize}

\textbf{Output Format}\\
Please predict this person's age. Please output a single number indicating the age highlighted with [] (e.g., [x]). Only output your answer in brackets.
\end{mdframed}

\phantomsection\label{prompt:bigfive-seq}
\paragraph{Task 4: Sequential Survey Response Prediction.}
\begin{mdframed}[style=promptstyle]
\textbf{System:} You are an expert in psychology. Given a subject's previous answers in a Big Five personality test (in the order they answered them), your task is to predict this subject's answer to the next question.

\medskip
\textbf{User:}\\
\textbf{Subject's Previous Answers}\\
The following items were rated on a five point scale where 1=Disagree, 2=Slightly Disagree, 3=Neutral, 4=Slightly Agree, 5=Agree.

\ph{previous\_answers}

\textbf{The Next Question}\\
\ph{next\_question}

\textbf{Output Format}\\
Please predict the subject's answer to the next question and highlight your prediction in [](such as [1],[2],[3],[4],or [5]). Only output your answer in brackets.
\end{mdframed}

\noindent
Items are numbered by original survey position (1--50); no item codes are shown.

\phantomsection\label{prompt:bigfive-acrossitem}
\paragraph{Task 5: Masked Survey Response Prediction.}
\begin{mdframed}[style=promptstyle]
\textbf{System:} You are an expert in psychology. Given a subject's answers to 9 questions measuring the same Big Five personality dimension, your task is to predict this subject's answer to the remaining question in that dimension.

\medskip
\textbf{User:}\\
\textbf{Subject's Answers (\ph{dimension\_name} Dimension)}\\
The following items were rated on a five point scale where 1=Disagree, 2=Slightly Disagree, 3=Neutral, 4=Slightly Agree, 5=Agree.

\ph{context\_answers}

\textbf{The Remaining Question}\\
\ph{target\_question}

\textbf{Output Format}\\
Please predict the subject's answer to the remaining question and highlight your prediction in [](such as [1],[2],[3],[4],or [5]). Only output your answer in brackets.
\end{mdframed}

\noindent
\ph{context\_answers} lists 9 items numbered 1--9 (no item codes).

\phantomsection\label{prompt:bigfive-leaveone}
\paragraph{Task 6: Across-Dimension Personality Prediction.}
\begin{mdframed}[style=promptstyle]
\textbf{System:} You are an expert in psychology. Given a person's BigFive personality test scores in four dimensions, your task is to predict the score in the remaining dimension.

\medskip
\textbf{User:}\\
\textbf{BigFive Dimensionality Scores}\\
Each dimensionality score ranges from 10 to 50, with 10 indicating the lowest score in that dimension and 50 indicating the highest score.

\ph{dimension\_scores}

\textbf{Output Format}\\
Based on the four dimension scores above, please predict this person's score in the *\ph{target\_name}* dimension. Please output a single number in the range from 10 to 50, highlighted in [] (e.g., [x]). Only output your answer in brackets.
\end{mdframed}

\noindent
\ph{dimension\_scores} lists 4 of the 5 dimensions (Openness, Conscientiousness, Extraversion, Agreeableness, Neuroticism) with their scores.

\phantomsection\label{prompt:sciworkflow}
\subsection*{E.\quad Scientific Workflow Prediction}

\paragraph{System prompt.}
\begin{mdframed}[style=promptstyle]
You are an expert in research tasked with generating detailed prompts for various aspects of academic research papers. Each task involves creating a specific type of prompt based on the provided information. Here are the definitions of each part you will work with:
\begin{itemize}[nosep]
\item \textbf{Context:} The status quo of related literature or reality which motivated this study. This could normally be a problem, a research question, or a research gap that has not been successfully addressed by previous work. This is anything that happened before this study.
\item \textbf{Key Idea:} The main intellectual merit of this paper, often in comparison to the context. This could normally be a novel idea or solution proposed in this paper that distinguishes it from what's already done in literature. This is proposed in this study.
\item \textbf{Method:} The specific research method that investigates and validates the key idea. This could be an experimental setup, a theoretical framework, or other necessary methodology to implement and/or evaluate the key idea. This is performed in this study.
\item \textbf{Outcome:} The factual statement about the study output. This could be the experiment results and any other measurable outcome that has occurred. It marks whether the key hypothesis is testified or not. This is produced in this study.
\item \textbf{Projected Impact:} The author-anticipated impact of the work on the field, and potential further research identified by the author that may improve or extend this study. This is anything being anticipated but has not happened yet.
\end{itemize}
\end{mdframed}

\paragraph{Task user prompt templates.}
The five workflow prediction tasks use the following user prompt templates:

\begin{mdframed}[style=promptstyle]
\textit{1. Idea Generation:}\\
Given the context: `\ph{context}', generate one key idea that could advance this area of study. Output only the key idea without explanation.

\medskip
\textit{2. Method Recommendation:}\\
Given the context: `\ph{context}' and the key idea: `\ph{key\_idea}', recommend one method to validate this idea. Output only the method without explanation.

\medskip
\textit{3. Outcome Prediction:}\\
Based on the context: `\ph{context}', the key idea: `\ph{key\_idea}', and the method: `\ph{method}', predict the outcome of this research. Output only the outcome without explanation.

\medskip
\textit{4. Impact Prediction:}\\
Based on the context: `\ph{context}', the key idea: `\ph{key\_idea}', the method: `\ph{method}', and the outcome: `\ph{outcome}', suggest the projected impact for this research. Output only the impact without explanation.

\medskip
\textit{5. Title Prediction:}\\
Given the context: `\ph{context}', the key idea: `\ph{key\_idea}', the method: `\ph{method}', the outcome: `\ph{outcome}', and the projected impact: `\ph{projected\_impact}', predict the title of this research paper. Output only the title without explanation.
\end{mdframed}

\phantomsection\label{prompt:ieo}
\subsection*{F.\quad IEO Economics Contest}

\begin{mdframed}[style=promptstyle]
\textbf{System:} You are an expert in economics. Answer the following multiple-choice question by selecting the correct option (A, B, C, or D). Provide only the letter of your answer.

\medskip
\textbf{User:} \ph{question}

A. \ph{option\_A}\\
B. \ph{option\_B}\\
C. \ph{option\_C}\\
D. \ph{option\_D}
\end{mdframed}

\end{small}

%% file: sections/appendix/asset-licenses.tex
\section{Licenses for Existing Assets}
\label{app:data-licenses}

\bench is constructed from the publicly available data sources described in Section~\ref{app:data-sources}. We summarize the licensing and terms-of-use status of each below.

\paragraph{MobLab economic-games data.} The MobLab game logs we use are released by \citet{mei2024turing} via the public repository \url{https://github.com/yutxie/ChatGPT-Behavioral} under the MIT License.

\paragraph{Big Five Personality Test.} The Big Five responses are obtained from the publicly available Kaggle dataset \url{https://www.kaggle.com/datasets/lucasgreenwell/ocean-five-factor-personality-test-responses}, which mirrors data distributed by the Open-Source Psychometrics Project (\url{https://openpsychometrics.org/_rawdata/}). The dataset is released under the Database Contents License (DbCL) v1.0 (\url{https://opendatacommons.org/licenses/dbcl/1-0/}).

\paragraph{Scientific articles (AER, NHB).} We use only the titles and abstracts of articles published in the \emph{American Economic Review} (American Economic Association) and \emph{Nature Human Behaviour} (Springer Nature). All articles are properly cited, and titles and abstracts are used for non-commercial research purposes consistent with standard fair-use practice for bibliographic data.

\paragraph{International Economics Olympiad.} Past IEO multiple-choice problems are publicly distributed for educational use (\url{https://ieo-official.org/}). We use them only for non-commercial research evaluation.